%% file: neurips_2026.tex
\documentclass{article}

\usepackage[preprint]{neurips_2026}

\usepackage[utf8]{inputenc}
\usepackage[T1]{fontenc}
\usepackage{xcolor}
\definecolor{deepblue}{RGB}{0, 70, 140}
\definecolor{tealblue}{RGB}{0, 130, 130}

\usepackage[colorlinks=true,
            linkcolor=deepblue,
            citecolor=deepblue,
            urlcolor=tealblue]{hyperref}
\usepackage{url}
\usepackage{graphicx}
\usepackage{booktabs}
\usepackage{amsfonts}
\usepackage{amsmath}
\usepackage{amssymb}
\usepackage{nicefrac}
\usepackage{microtype}
\usepackage{xcolor}
\usepackage{multirow}
\usepackage{xspace}
\usepackage{bm}
\usepackage[capitalise,nameinlink,noabbrev]{cleveref}

\graphicspath{{figures/}}

\newcommand{\mfc}{MobileFetalCLIP\xspace}

\newcommand{\lfeat}{\lambda_{\mathrm{feat}}}
\newcommand{\eg}{e.g.}

\newcommand{\etal}{et al.}
\providecommand{\keywords}[1]{}

\newcommand{\x}{\bm{\mathrm{x}}}

\newcommand{\fplane}{F1$_{\mathrm{5Plane}}$~}
\newcommand{\fbrain}{F1$_{\mathrm{3Brain}}$~}
\newcommand{\fplanebrain}{F1$_{\mathrm{Pl\_Br}}$~}
\newcommand{\vrhc}{VR$_{\mathrm{HC18}}$~}

\title{DARK: Diagonal-Anchored Repulsive Knowledge Distillation for Vision-Language Models under Extreme Compression}

\author{%
  Numan Saeed\thanks{Corresponding author: \texttt{numan.saeed@mbzuai.ac.ae}},
  Asif Hanif,
  Fadillah Adamsyah Maani,
  Hussain Alasmawi\\
  \textbf{Mohammad Yaqub}\\
  Computer Vision, Mohamed Bin Zayed University of Artificial Intelligence, Abu Dhabi, UAE\\
  \parbox{\linewidth}{\centering\footnotesize\ttfamily
    \{numan.saeed,\allowbreak\ asif.hanif,\allowbreak\ fadillah.maani,\allowbreak\ hussain.alasmawi,\\
    mohammad.yaqub\}@mbzuai.ac.ae
  }
}

\begin{document}

\maketitle

\begin{abstract}
Compressing vision-language models for on-device deployment is increasingly important in clinical settings, but knowledge distillation (KD) degrades sharply when the teacher-student capacity gap spans an order of magnitude or more. We argue that, under such gaps, strict imitation of the teacher is a poor objective: much of the teacher's pairwise similarity structure reflects its own architectural biases rather than information a compact student can efficiently represent. We propose \textbf{Diagonal-Anchored Repulsive Knowledge Distillation (DARK)}, a contrastive KD framework that decomposes the distillation loss into a diagonal term (matched image-text pairs) and an off-diagonal term (non-target similarities). The diagonal term anchors matched-pair alignment throughout training; the off-diagonal term is annealed from positive to negative weighting, transitioning the student from imitating to \emph{repelling} the teacher's non-target similarity structure. We instantiate DARK by distilling FetalCLIP, a 427M-parameter fetal ultrasound vision-language model, into \textbf{MobileFetalCLIP}, a 75M-parameter student model with a $26\times$ smaller visual encoder, running in 1.6\,ms on an iPhone~16~Pro. The student matches or exceeds its teacher on three zero-shot benchmarks, including HC18 biometry validity (88.6\% vs.\ 83.5\%) and brain sub-plane F1 (0.784 vs.\ 0.702). Embedding-geometry and logit analyses show that DARK induces \emph{structured decorrelation}: the student preserves teacher-aligned per-image confidence while diverging from inherited inter-class confusion, suggesting that controlled repulsion can be more efficient than imitation under extreme compression.

\end{abstract}

\section{Introduction}
\label{sec:intro}

Vision-language models have become a versatile foundation for transferable visual representations across natural and specialized domains. Their deployment in resource-constrained environments, however, remains a binding constraint on their practical impact. This tension is particularly acute in clinical imaging: domain-specialized models such as FetalCLIP~\cite{fetalclip2025} achieve strong zero-shot performance on fetal ultrasound tasks that could substantially expand access to prenatal care in low-resource settings~\cite{stewart2020ultrasound,pokaprakarn2022gestational,gomes2022mobile}, yet their hundreds of millions of parameters preclude deployment on the handheld systems that would carry them into those settings. Bridging this gap demands order-of-magnitude compression of the visual encoder while preserving zero-shot ability, a regime in which the standard compression tool, knowledge distillation, becomes unreliable.

Knowledge distillation~\cite{hinton2015kd} is the natural mechanism for compressing such models, transferring information from a large teacher to a smaller student. For CLIP-style architectures, recent work shows that zero-shot performance can be preserved when teacher and student have comparable capacity: TinyCLIP~\cite{tinyclip2023} relies on weight inheritance between architecturally compatible models, and CLIP-KD~\cite{clipkd2024} systematically benchmarks logit- and feature-level objectives at moderate compression ratios. Neither method targets the order-of-magnitude regime our setting requires. Prior analyses establish that distillation performance degrades as the capacity gap widens~\cite{cho2019efficacy,mirzadeh2020takd,stanton2021does}, and that strict imitation objectives can become ineffective when the teacher is substantially stronger than the student~\cite{huang2022dist,sun2024logstd}. The mechanism behind this degradation, and the question of what the student should attempt to inherit when full imitation is infeasible, remain open.

We hypothesize that the difficulty is not simply capacity but what the student is asked to imitate. The teacher's $N \times N$ batch similarity matrix carries two qualitatively different signals: the diagonal encodes matched image--caption alignment, a property every vision--language model must satisfy, while the off-diagonal encodes the teacher's specific pattern of inter-class similarity, which reflects its architectural priors as much as the data. A ViT-L/14 with global self-attention can support a richly distributed pattern of inter-class similarities that a $26\times$ smaller convolution--attention hybrid cannot, so asking the student to reproduce this off-diagonal structure misallocates limited capacity. Our embedding analysis (Section~\ref{sec:feature_space}) confirms this: under standard distillation, the student's similarity structure tracks the teacher's at the cost of weaker downstream cluster geometry. This motivates separating the two signals and treating them asymmetrically.


Building on this hypothesis, we propose \textbf{Diagonal-Anchored Repulsive Knowledge Distillation (DARK)}, which decomposes the symmetric cross-entropy distillation loss into a diagonal term (matched image--caption alignment) and an off-diagonal term (non-target similarities), and treats them asymmetrically. The diagonal term is anchored at a fixed positive weight throughout training; the off-diagonal term is scaled by a coefficient $\beta(t)$ that decays linearly from a positive starting value into negative territory. Once $\beta(t) < 0$, the off-diagonal gradient inverts and the student transitions from imitating to actively repelling the teacher's non-target similarity structure, the \emph{repulsive regime}.


We instantiate DARK by distilling FetalCLIP~\cite{fetalclip2025}, a 427M-parameter fetal-ultrasound vision-language model with a ViT-L/14 visual encoder, into \textbf{MobileFetalCLIP}, a 75M-parameter student built around an 11.4M-parameter FastViT~\cite{vasu2023fastvit} visual encoder, a $26\times$ reduction in visual-encoder parameters. Despite this gap, MobileFetalCLIP matches or exceeds the teacher across three zero-shot benchmarks: it surpasses the teacher on HC18 biometry validity (88.6\% vs.\ 83.5\%) and brain sub-plane classification F1 (0.784 vs.\ 0.702), while remaining competitive on five-plane classification (0.946 vs.\ 0.973). The visual encoder runs in 1.6\,ms on an iPhone~16~Pro, well above the throughput required for real-time clinical use.

To understand why the student can exceed its teacher, we analyze the geometry of the learned representations and the distribution of the student's logits, finding that DARK preserves the teacher's per-image confidence while decorrelating its inter-image similarity structure, an effect we term \emph{structured decorrelation}. Our \textbf{contributions} are as follows:


\begin{itemize}
    \item \textbf{A contrastive distillation framework for extreme compression.} We introduce \textbf{DARK}, which decomposes the contrastive KD objective into diagonal and off-diagonal terms and anneals the off-diagonal weight from positive to negative, transitioning the student from imitating to repelling the teacher's non-target similarity structure.
    \item \textbf{An empirical instantiation in fetal ultrasound.} We apply DARK to compress FetalCLIP into \textbf{MobileFetalCLIP}, a model with a $26\times$ smaller visual encoder that matches or exceeds its teacher on three zero-shot benchmarks and runs in 1.6\,ms on an iPhone~16~Pro. We release the model and training framework to support mobile medical AI.
    \item \textbf{Analysis of why repulsion helps.} We show that DARK induces \emph{structured decorrelation}: the student preserves matched-pair confidence inherited from the teacher while diverging from its inter-class similarity pattern, suggesting that selectively repelling, rather than imitating, the teacher's non-target structure can improve generalization under extreme compression.
\end{itemize}

\section{Related Work}
\label{sec:related}

\noindent\textbf{Knowledge Distillation under Capacity Constraints.}
Knowledge distillation (KD)~\cite{hinton2015kd} transfers information from a large teacher to a smaller student, with extensions incorporating intermediate feature matching (FitNets~\cite{romero2015fitnets}), relational structure (RKD~\cite{park2019rkd}), and contrastive objectives (CRD~\cite{tian2020crd}). For vision-language models, CLIP-KD~\cite{clipkd2024} explores logit- and feature-based distillation, while TinyCLIP~\cite{tinyclip2023} relies on weight inheritance, which requires architectural compatibility between teacher and student.

A key challenge arises under large capacity gaps. Distillation performance degrades as the gap widens~\cite{cho2019efficacy,mirzadeh2020takd,stanton2021does}, and strict matching objectives can become ineffective or unstable when the teacher is substantially stronger than the student~\cite{huang2022dist,sun2024logstd}. Existing mitigations include intermediate teacher assistants~\cite{mirzadeh2020takd}, softened or relaxed matching objectives~\cite{huang2022dist,sun2024logstd}, and architectural alignment constraints~\cite{tinyclip2023}. While Born-Again Networks~\cite{furlanello2018born} have shown that students can occasionally surpass their teachers, this phenomenon has been studied primarily in the matched-architecture, matched-capacity regime; the mechanisms that enable it under order-of-magnitude capacity gaps remain underexplored.

The closest method is Decoupled Knowledge Distillation (DKD)~\cite{zhao2022dkd}, which separates the classification distillation loss into target and non-target components weighted independently. We extend this decomposition to the contrastive $N \times N$ similarity matrix and, crucially, anneal the off-diagonal weight from positive to negative,a sign-flipped regime not considered by DKD or, to our knowledge, by any prior contrastive distillation method, which lets the student depart from inherited inter-class relationships rather than approximate them.

\noindent\textbf{Regularisation and Representation Decorrelation.}
Barlow Twins~\citep{zbontar2021barlow} and the alignment/uniformity framework of \citet{wang2020uniformity} motivate the geometric analyses we use in \cref{sec:feature_space} (uniformity, participation-ratio dimensionality). DARK is also distinct from undirected confidence regularisation: \citet{pereyra2017penalizing} penalize overconfidence uniformly, whereas DARK's repulsive signal is teacher-derived and directional, identifying which non-matching pairs to separate. Replacing this directional signal with a uniform entropy penalty yields no improvement over CLIP-only training (\cref{sec:ablation}).

\noindent\textbf{Vision-Language Pretraining and Medical Specialization.}
CLIP~\cite{radford2021clip} introduced contrastive image--text pretraining for transferable zero-shot representations, with extensions improving objectives (SigLIP~\cite{zhai2023siglip}), scaling (OpenCLIP~\cite{ilharco2021openclip,cherti2023reproducible}), and efficiency (MobileCLIP~\cite{mobileclip2024,mobileclip2_apple}). Our student adopts a mobile-scale design based on MobileCLIP, using a FastViT~\cite{fastvit2023} encoder for favorable accuracy--efficiency trade-offs. In medical imaging, MedCLIP~\cite{medclip2022}, BiomedCLIP~\cite{biomedclip2024}, and UniMed-CLIP~\cite{unimedclip2024} extend vision--language pretraining to biomedical domains, while CheXzero~\cite{chexzero2022} demonstrates expert-level zero-shot performance via modality-specific training. FetalCLIP~\cite{fetalclip2025}, our teacher, specializes this paradigm to fetal ultrasound, but its 427M parameters limit on-device use, motivating our compression approach.

\vspace{-0.1em}


\begin{figure}
    \centering    \includegraphics[width=1.0\linewidth]{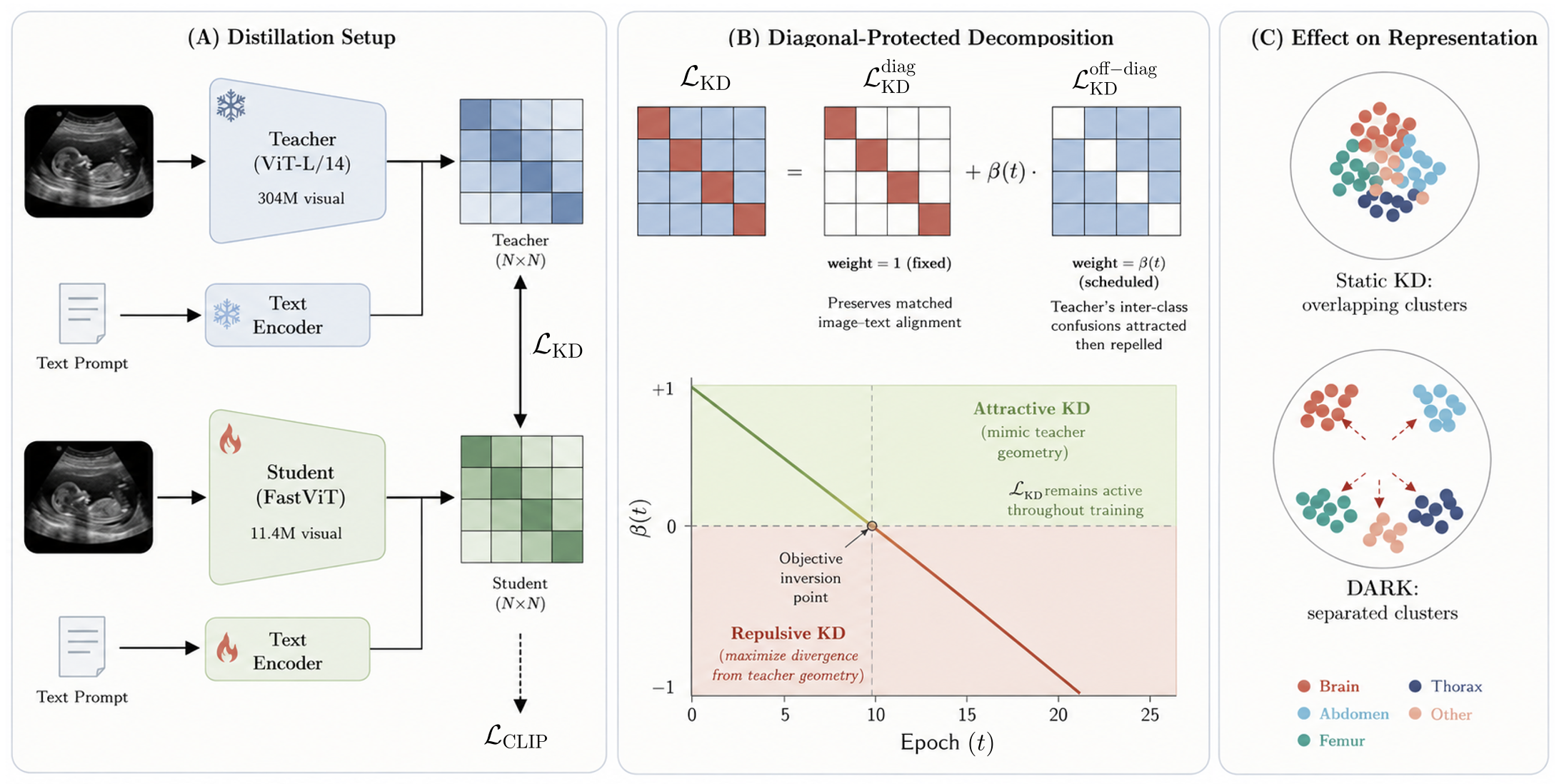}
    \caption{Overview of the DARK framework. (A) Distillation setup: a frozen FetalCLIP teacher (ViT-L/14, 304M visual params) produces an $N\times N$ similarity matrix; a lightweight FastViT student (11.4M visual params) is trained via $\mathcal{L}_{\mathrm{CLIP}}$ and $\mathcal{L}_{\mathrm{KD}}$. (B) Attraction-to-repulsion dynamics: the off-diagonal weight $\beta(t)$ linearly decays from positive $\beta_{\mathrm{start}}$ into a negative value $\beta_{\mathrm{stop}}$; the weight for $\mathcal{L}_{\mathrm{diag}}$ term remains fixed, preserving matched pair alignment throughout training. (C) Outcome: DARK produces structured decorrelation, resulting in better cluster separation and a higher \vrhc and \fbrain with $26\times$ fewer visual parameters.}
    \label{fig:method-diagram}
\end{figure}

\section{Method}
\label{sec:method}

We present \textbf{Diagonal-Anchored Repulsive Knowledge Distillation (DARK)}, a contrastive distillation framework for compressing vision-language models under large teacher-student capacity gaps. DARK rests on two observations. First, the contrastive distillation loss admits a natural decomposition into a \emph{diagonal} term, transferring matched image-caption alignment, and an off-diagonal term, transferring inter-class similarity structure. Second, these two terms encode qualitatively different information: the diagonal term expresses a property every vision-language model must satisfy, while the off-diagonal term reflects the teacher's specific pattern of inter-class confusions, which depends on its architectural priors as much as on the data. Under extreme compression, faithfully imitating the off-diagonal misallocates limited student capacity to relations the student's architecture is poorly suited to represent. DARK therefore \emph{anchors} the diagonal term and \emph{anneals} the off-diagonal term from positive (attractive) to negative (repulsive) over training. An overview is shown in \cref{fig:method-diagram}.

\subsection{Preliminaries}
\label{sec:prelim}

\noindent\textbf{Contrastive image-text pretraining.}
Let $\{(\x_i, t_i)\}_{i=1}^N$ be a batch of $N$ image-text pairs. CLIP~\cite{radford2021clip} learns image and text encoders $f_I$, $f_T$ such that paired embeddings have higher cosine similarity than non-paired ones. Writing $s_{ij} = f_I(\x_i)^{\!\top} f_T(t_j)$ for the (normalized) similarity, the symmetric InfoNCE objective is
\begin{equation}
  \mathcal{L}_\mathrm{CLIP}
  = -\frac{1}{2N}\sum_{i=1}^{N}\!\left[
    \log\frac{e^{s_{ii}/\tau}}{\sum_{j} e^{s_{ij}/\tau}}
    \;+\;
    \log\frac{e^{s_{ii}/\tau}}{\sum_{j} e^{s_{ji}/\tau}}
  \right],
  \label{eq:clip}
\end{equation}
where $\tau$ is a temperature.

\noindent\textbf{Logit-level distillation.}
We adopt the logit-matching framework of CLIP-KD~\cite{clipkd2024}: a frozen teacher and a trainable student each produce an $N{\times}N$ similarity matrix $S^{\mathrm{T}}$, $S^{\mathrm{S}}$ on the same batch. Their row-wise softmax distributions are
\begin{equation}
  p^{\mathrm{T}}_i = \mathrm{softmax}\!\bigl(S^{\mathrm{T}}_{i,:}/\tau_\mathrm{KD}\bigr),
  \qquad
  q^{\mathrm{S}}_i = \mathrm{softmax}\!\bigl(S^{\mathrm{S}}_{i,:}\bigr),
  \label{eq:softmax}
\end{equation}
where $\tau_\mathrm{KD}$ is a fixed KD temperature applied only to the
teacher.
The symmetric distillation loss averages the image-to-text $(I\to T)$ and text-to-image $(T\to I)$ directions of the similarity matrix,
\begin{equation}
  \mathcal{L}_\mathrm{KD}
  = \tfrac{1}{2}\!\left[\,
      \mathcal{H}\!\left(p^{\mathrm{T}}, q^{\mathrm{S}}\right)_{I\to T}
      + \mathcal{H}\!\left(p^{\mathrm{T}}, q^{\mathrm{S}}\right)_{T\to I}
   \,\right],
  \label{eq:kd_main}
\end{equation}
where $\mathcal{H}(p,q)=-\sum_j p_j\log q_j$ is the cross-entropy, which is equivalent to KL divergence $\mathrm{KL}\!\left(p^{\mathrm{T}}\,\|\,q^{\mathrm{S}}\right)$ up to additive teacher-entropy constants. Note that we apply $\tau_\text{KD}$ only to the teacher and leave the student's logits unscaled, omitting the standard $\tau_\text{KD}^2$ prefactor of Hinton-style KD. In the contrastive setting the student's learned logit scale already calibrates the sharpness of its similarity distribution, so an additional $\tau_\text{KD}$ on the student would interfere with this learned calibration.

\subsection{Diagonal-Off-Diagonal Decomposition of Distillation Loss}
\label{sec:decomp}

For row $i$, the cross-entropy in \cref{eq:kd_main} splits according to whether the column index matches the row index:
\begin{equation}
\mathcal{H}\!\left(p^{\mathrm{T}}_i, q^{\mathrm{S}}_i\right)
  = \underbrace{-\,p^{\mathrm{T}}_{ii}\log q^{\mathrm{S}}_{ii}\vphantom{\sum_{j\neq i}}}_{\text{matched (diagonal)}}
    \;+\;
    \underbrace{-\!\sum_{ j\neq i} p^{\mathrm{T}}_{ij}\log q^{\mathrm{S}}_{ij}}_{\text{non-target (off-diagonal)}}.
    \label{eq:cross-entropy-split}
\end{equation}

While Equation~(\ref{eq:cross-entropy-split}) is a trivial rewrite of the cross-entropy, it isolates two distinct signals carried by $p^T_i$: the matched-pair confidence $p^T_{ii}$, encoding the teacher's belief that image $i$ aligns with caption $i$, and the non-target distribution $\{p^T_{ij}\}_{j \neq i}$, encoding the teacher's specific pattern of inter-class similarity. DARK exploits this separation by treating the two signals asymmetrically. Averaging over the batch and over the two contrastive directions yields
\begin{equation}
  \mathcal{L}_\mathrm{KD}
  = \mathcal{L}_\mathrm{KD}^{\mathrm{diag}} + \mathcal{L}_\mathrm{KD}^{\mathrm{off}\text{-}\mathrm{diag}},
  \label{eq:decomp}
\end{equation}
where $\mathcal{L}_\mathrm{KD}^{\mathrm{diag}}$ governs transfer of confidence on \emph{matched} image-caption pairs, and $\mathcal{L}_\mathrm{KD}^{\mathrm{off}\text{-}\mathrm{diag}}$ governs transfer of the \emph{inter-class} similarity structure. \Cref{eq:decomp} is the contrastive analogue of the target/non-target split introduced for classification by Decoupled KD~\cite{zhao2022dkd}, generalised here to the $N{\times}N$ similarity matrix.

The two components play distinct and asymmetric roles. The diagonal term enforces correct image–text alignment, a fundamental requirement for the student to function as a vision–language model. In contrast, the off-diagonal term captures the teacher’s non-target similarity structure, including inter-class confusions. Under a large architectural gap, a substantial portion of this structure reflects the inductive biases of the teacher (e.g., global self-attention) rather than intrinsic ambiguity in the data. Enforcing strict alignment with these off-diagonal patterns can therefore lead a compact student to allocate capacity to relations that are not naturally supported by its architecture; we provide empirical support for this in  \cref{sec:feature_space}, where DARK substantially improves cluster geometry over static KD under the same architecture and training data.

\subsection{Annealed Attraction-to-Repulsion Schedule}
\label{sec:schedule}

Our proposed method DARK exploits the asymmetry in Equation~\ref{eq:decomp} by treating the two components differently. The diagonal term is assigned a fixed weight of $1$ throughout training, while the off-diagonal term is scaled by an epoch-dependent coefficient $\beta(t)$. Starting from an initial value $\beta_{\mathrm{start}}$, it decays linearly over training and reaches a final value $\beta_{\mathrm{stop}}$ after $E$ epochs. Formally, at epoch\footnote[2]{We express the schedule in epoch-units for clarity; in practice $t$ indexes optimization steps and $\beta(t)$ is updated step-wise based on overall training progress.} $t$:
\begin{equation}
\beta(t) = \beta_{\mathrm{start}} + \frac{t}{E}\bigl(\beta_{\mathrm{stop}} - \beta_{\mathrm{start}}\bigr).
\label{eq:schedule}
\end{equation}
Setting $\beta_{\mathrm{stop}} \geq 0$ results in a standard decaying-KD schedule. Allowing $\beta_{\mathrm{stop}} < 0$ enables $\beta(t)$ to cross zero at
$ t = (\beta_{\mathrm{start}}\cdot E)/(\beta_{\mathrm{start}} -\beta_{\mathrm{stop}})$
after which $\beta(t) < 0$ and the gradient of $\beta(t)\cdot\mathcal{L}_\mathrm{KD}^{\mathrm{off}\text{-}\mathrm{diag}}$ inverts: rather than minimising the divergence between the student and teacher off-diagonal distributions, the objective maximises it, actively repelling the student from the teacher's non-target similarity structure. We refer to this as the \emph{repulsive regime}. This annealed schedule creates a smooth transition from imitation to repulsion and, as shown empirically in the results, improves student performance under large capacity gaps. Training therefore proceeds through three phases (\cref{fig:method-diagram}). In the \textbf{attractive phase} ($\beta(t) > 0$), the student absorbs the teacher's relational structure. As $\beta(t)$ approaches zero, the training enters a \textbf{transition phase}, in which the off-diagonal KD term becomes negligible, and optimization is dominated by $\mathcal{L}_\mathrm{CLIP}$. Once $\beta(t) < 0$, the model enters the \textbf{repulsive phase}, where the student is encouraged to diverge from the teacher's off-diagonal similarities, while the always-positive $\mathcal{L}_\mathrm{KD}^{\mathrm{diag}}$ and $\mathcal{L}_\mathrm{CLIP}$ preserve matched-pair alignment.

\subsection{DARK Training Objective}
\label{sec:objective}

Combining \cref{eq:clip,eq:decomp,eq:schedule}, the final objective at training step $t$ is
\begin{equation}
  \mathcal{L}(t)
  \;=\;
  \mathcal{L}_\mathrm{CLIP}
  \;+\;
  \mathcal{L}_\mathrm{KD}^{\mathrm{diag}}
  \;+\;
  \beta(t)\cdot\mathcal{L}_\mathrm{KD}^{\mathrm{off}\text{-}\mathrm{diag}}.
  \label{eq:total_loss}
\end{equation}
Here, $\mathcal{L}_\mathrm{CLIP}$ is the standard contrastive loss applied to the student, while $\mathcal{L}_\mathrm{KD}^{\mathrm{diag}}$ and $\mathcal{L}_\mathrm{KD}^{\mathrm{off}\text{-}\mathrm{diag}}$ are as defined in \Cref{eq:decomp}. This formulation fixes the diagonal term and modulates the off-diagonal term via the epoch-dependent coefficient $\beta(t)$, yielding a smooth transition from standard distillation ($\beta(t) > 0$) to a repulsive regime ($\beta(t) < 0$) while preserving matched image--text alignment. We allow $\beta_{\mathrm{start}} > 1$ to emphasize non-target structure during the early (attractive) phase. ~\cite{zhao2022dkd} observe that the non-target component benefits from stronger weighting in the classification setting; we find that a moderate $\beta_{\mathrm{start}}$ = 2 suffices in our contrastive setting (\cref{sec:ablation}). 

Conceptually, DARK occupies an intermediate position between standard KD (which pulls the student toward the full teacher distribution) and uniform confidence regularisation (which pushes toward uniform); see Section~\ref{sec:related} for an extended discussion.

\section{Experiments and Results}
\label{sec:experiments}

\subsection{Implementation Setup}
\label{sec:setup}

\paragraph{Model Architectures.}
We investigated existing CLIP-based models. FetalCLIP~\cite{fetalclip2025} was trained on a large corpus of fetal ultrasound image-text pairs and has demonstrated strong performance across various fetal ultrasound tasks, making it a suitable teacher model. During distillation, we freeze FetalCLIP's vision encoder (ViT-L/14, 304M parameters).
The student is MobileFetalCLIP, built on a FastViT image encoder~\cite{mobileclip2024,fastvit2023} (11.4M visual parameters) with a 4‑layer Transformer text encoder; both produce 512‑dimensional embeddings.

\noindent\textbf{Pretraining Dataset.}
We curated a total of 246{,}349 fetal ultrasound image-caption pairs from a tertiary hospital and supplemented them with expert-annotated textbook image-caption pairs, as described in supplementary~\S\ref{sub:data_pipeline}.
To ensure that student and teacher logit matrices correspond to the same augmented input, we couple their data augmentations, sharing affine and color-jitter parameters within each forward pass.

\noindent\textbf{Evaluation Dataset.}
We evaluate zero-shot performance on two public benchmarks. \textit{Planes DB}~\cite{fetalplanesdb2020} comprises 12,400 fetal ultrasound images from 1,792 patients across two hospitals; we use 8,187 images for 5-plane classification (abdomen, brain, femur, thorax, cervix; excluding ``Other'') and 2,949 brain images for 3-class sub-plane classification (transthalamic, transcerebellum, transventricular), following FetalCLIP~\cite{fetalclip2025} by evaluating on the full labeled set without an added split. \textit{HC18}~\cite{hc18dataset2018} contains 999 head-circumference images; we retain 814 with plausible HC (100--342~mm; 14--40 weeks Gestational Age (GA)) and report \emph{validity rate} as in FetalCLIP~\cite{fetalclip2025}: GA is predicted via similarity to GA-specific text prompts, and a prediction is \emph{valid} if the true HC lies within the 2.5th--97.5th percentile of WHO growth charts~\cite{kiserud2017who} for the predicted GA.

\noindent\textbf{Optimisation.}
We train for $E=20$ epochs with effective batch size $1{,}024$ and KD temperature $\tau_{\mathrm{KD}}=5.0$ (as in CLIP-KD~\cite{clipkd2024}); the higher temperature spreads off-diagonal mass and strengthens the repulsive signal once $\beta(t)$ becomes negative. Unless noted, $\beta_{\mathrm{start}}=2$ and $\beta_{\mathrm{stop}}=-0.8$, yielding $\sim$70\% attractive and 30\% repulsive phases. Full hyperparameters, optimizer/scheduler, mixed-precision, and data pipeline details are in supplementary~\S\ref{app:impl}.

\noindent\textbf{Baselines.}
We compare with CLIP (ViT-L/14)~\cite{radford2021clip}, BiomedCLIP~\cite{biomedclip2024}, UniMed-CLIP~\cite{unimedclip2024}, SonoNet~\cite{sononet2017}, and the FetalCLIP teacher~\cite{fetalclip2025}, using reported results from Maani \etal~\cite{fetalclip2025}. Our primary distillation baseline is \textbf{static logit KD} ($\lambda{=}1.0$) with the CLIP-KD~\cite{clipkd2024} symmetric cross-entropy objective. TinyCLIP-style weight inheritance~\cite{tinyclip2023} is not applicable due to architectural mismatch (FastViT student vs.\ ViT-L/14 teacher).

\noindent\textbf{Metrics.}
All metrics are evaluated in the zero-shot setting. We report the HC18 validity rate (VR$_{\mathrm{HC18}}$), macro-F1 for 5-plane classification (F1$_{\mathrm{5Plane}}$), and macro-F1 for 3-class brain sub-plane classification (F1$_{\mathrm{3Brain}}$). To summarize overall classification performance, we define a class-balanced aggregate $
\mathrm{F1}_{\mathrm{Pl\text{\_}Br}} = ({5\times \mathrm{F1}_{\mathrm{5Plane}} + 3\times \mathrm{F1}_{\mathrm{3Brain}}})/{8},$
which weights each task by its number of classes. We also report ${\mathrm{Composite~Score}} = (\mathrm{F1}_{\mathrm{Pl\text{\_}Br}} + \mathrm{VR}_\mathrm{HC18})/2$. More details about the evaluation metric can be found in supplementary \S\ref{sub:eval_metric}.

\subsection{Results and Discussion}
\label{sec:main_results}

\Cref{tab:sota} summarizes zero-shot performance. Without distillation, the student achieves VR$_{\mathrm{HC18}}$ of 0.713 and F1$_{\mathrm{5Plane}}$ of 0.889, indicating that contrastive pretraining alone provides a reasonable initialization. Static Logit KD (CLIP-KD~\cite{clipkd2024}) yields clear gains (VR$_{\mathrm{HC18}}$: 0.794, F1$_{\mathrm{5Plane}}$: 0.946), confirming effective knowledge transfer from the teacher; however, a gap to the teacher remains (0.794 vs.\ 0.835 in VR$_{\mathrm{HC18}}$), consistent with capacity mismatch effects~\cite{cho2019efficacy}. Applying a time-varying and linearly decaying weight $\beta(t)$ to the full $\mathcal{L}_{\mathrm{KD}}$ loss (Coupled Repulsive KD) yields further gains (VR$_{\mathrm{HC18}}$: 0.844, F1$_{\mathrm{3Brain}}$: 0.763), indicating that gradually relaxing purely attractive distillation improves transfer. Our proposed \textsc{DARK} framework delivers the strongest results across metrics. With $26\times$ fewer visual parameters, \mfc surpasses the FetalCLIP teacher on VR$_{\mathrm{HC18}}$ (0.886 vs.\ 0.835) and F1$_{\mathrm{3Brain}}$ (0.784 vs.\ 0.702), while remaining competitive on F1$_{\mathrm{5Plane}}$ (0.946 vs.\ 0.973). This yields the best overall score (0.886). The improvement from the CLIP-KD baseline to \textsc{DARK} is primarily attributable to the revised distillation strategy (attractive-to-repulsive weight decay on off-diagonal term of KD loss), as architecture and training pipeline remain the same.

\input{table_sota}

\input{table_ablation.tex}

\noindent\textbf{Inference Efficiency.}
\Cref{tab:efficiency_main} compares the visual encoder's computational cost.
The student requires 32$\times$ fewer multiply-accumulate operations (MACs) and 26$\times$ fewer parameters than the teacher. On an iPhone\,16\,Pro the encoder runs in 1.6\,ms (24$\times$ faster than the teacher's 37.6\,ms), corresponding to over 600 frames per second (fps), well beyond the 30--60\,fps typical of diagnostic ultrasound. This throughput headroom means the encoder can be embedded in an on-device assistive pipeline for real-time standard-plane identification without interfering with the clinical scanning workflow.

\noindent\textbf{Training Dynamics.}
\Cref{fig:dynamics} shows training dynamics across KD variants. Both repulsive approaches, Coupled Repulsive KD and our \textsc{DARK} (Diagonal-Anchored Repulsive KD), exhibit a characteristic late-stage surge in zero-shot performance once the KD weight $\beta(t)$ becomes negative. This transition flips the KD objective from attraction to repulsion on non-target pairs, discouraging the student from inheriting the teacher’s inter-class confusion and promoting more discriminative representations. While Coupled Repulsive KD already improves over static KD, \textsc{DARK} consistently achieves stronger gains, reaching the highest $\textsc{Composite~Score}$ of 0.886 (vs.\ 0.853 for the teacher), highlighting the benefit of gradually structuring the repulsive signal. Extended dynamics are provided in supplementary Figures~\ref{fig:dynamics_all}--\ref{fig:composite_trajectories}.

\begin{table}[t]
\centering
\begin{minipage}[t]{0.48\linewidth}
    \centering
    \caption{Inference efficiency: visual-encoder parameters, GMACs,
      and on-device encoder latency (CoreML, fp16, batch\,1).
      Measured on two iPhones to show generational consistency.}
    \label{tab:efficiency_main}
    \centering
    \scriptsize
    \setlength{\tabcolsep}{1pt}
    \begin{tabular}{@{}lrrrr@{}}
    \toprule
     & & & \multicolumn{2}{c}{iPhone Latency (ms)} \\
    \cmidrule(l){4-5}
    \textsc{Model} & \textsc{Params.} & \textsc{GMACs} & 16\,\textsc{Pro} & 17\,\textsc{Pro} \\
    \midrule
    FetalCLIP & 304M & 38.9 & 37.6 & 31.9 \\
    \mfc & 11.4M {\scriptsize(26$\times\!\downarrow$)} & 1.2 {\scriptsize(32$\times\!\downarrow$)} & 1.6 {\scriptsize(24$\times\!\downarrow$)} & 1.4 {\scriptsize(23$\times\!\downarrow$)} \\
    \bottomrule
    \end{tabular}
\end{minipage}\hfill
\begin{minipage}[t]{0.48\linewidth}
      \caption{%
        Embedding geometry on Planes DB (5-plane, 8{,}187 images).
        $d_\text{eff}$: effective dimensionality (participation ratio).
        Rank$_{95}$: singular values for 95\% variance.
      }
      \label{tab:geometry}
      \centering
      \scriptsize
      \setlength{\tabcolsep}{1pt}
      \begin{tabular}{@{}lcccccc@{}}
        \toprule
        \textsc{Method} & $d_\text{eff}$ & \textsc{Rank}$_{95}$ & \textsc{Silh.}\,$\uparrow$ & \textsc{Intra}\,$\uparrow$ & \textsc{Inter}\,$\downarrow$ & \textsc{Unif.}\,$\downarrow$ \\
        \midrule
        Static KD & 8.0 & 77 & 0.375 & 0.712 & 0.445 & $-$1.662 \\
        Conf.\ Penalty & 9.0 & 84 & 0.406 & 0.693 & 0.389 & $-$1.811 \\
        Coupled R. KD & 6.4 & 50 & 0.509 & 0.645 & $\mathbf{0.010}$ & $-$2.231 \\
        DARK (ours) & \textbf{10.0} & 74 & \textbf{0.525} & 0.623 & 0.076 & $\mathbf{-2.308}$ \\
        \bottomrule
      \end{tabular}
\end{minipage}
\end{table}

\subsection{Ablative Analysis}
\label{sec:ablation}

\Cref{tab:ablation} presents the full ablation. The results reveal a clear progression:

\noindent\textbf{Static KD helps but plateaus.} Static KD ($\mathcal{L}_{\mathrm{CLIP}}+\lambda\cdot\mathcal{L}_{\mathrm{KD}}$) with $\lambda{=}1$ improves \fplane (0.889$\to$0.946) and \vrhc (0.713$\to$0.794), but remains below the teacher (0.835) and yields minimal gains on \fbrain (0.712$\to$0.715). \textbf{Weakening or removing KD is ineffective.} Introducing decay in standard KD with $\lambda$ varying from $1\to0.1$ reduces \vrhc (0.746), and full decay ($\lambda:1\to0$) performs worse (0.731), indicating the need for sustained teacher guidance. \textbf{Feature KD is detrimental.} Adding feature alignment~\cite{clipkd2024} to Static KD degrades both \vrhc (0.794$\to$0.759) and \fbrain (0.715$\to$0.664), suggesting that pointwise mimicry is harmful under large capacity gaps. \textbf{Unstructured regularization is insufficient.}  Confidence penalty method \cite{pereyra2017penalizing} ($\varepsilon{=}0.1$) also performs poorly (\vrhc 0.749, \fbrain 0.680), showing that entropy alone cannot replace structured distillation. \textbf{Coupled Repulsive KD overcomes the gap.} Allowing $\beta(t){<}0$ in $(\mathcal{L}_{\mathrm{CLIP}}+\beta(t)\cdot\mathcal{L}_{\mathrm{KD}})$, termed as Coupled Repulsive KD, boosts \vrhc to 0.844 (surpassing the teacher) and improves \fbrain to 0.763. \textbf{\textsc{DARK} performs best.} By preserving the diagonal KD and gradually transitioning from attraction to repulsion of off-diagonal KD, \textsc{DARK} further improves \vrhc to 0.886 and \fbrain to 0.784, outperforming both the teacher and coupled repulsion KD. The gain over coupled KD ($+$4.2\% \vrhc, $+$2.1\% \fbrain) highlights the importance of late stage off-diagonal repulsion, particularly for retrieval tasks. Extreme schedules (stronger amplification i.e. $\beta_{\mathrm{start}}\!\in\!\{4,8\}$ or weaker repulsion i.e.  $\beta_{\mathrm{stop}}\!\in\!\{-0.4,-0.5\}$ in DARK training objective degrade performance, indicating that both decomposition and repulsion strength are critical. We also perform \textbf{linear probing} evaluation and provide results in supplementary \S\ref{app:linear_probe}.

\subsection{Feature Space Analysis}
\label{sec:feature_space}

\begin{figure}[t]
    \centering
    \begin{minipage}[t]{0.49\linewidth}
      \centering
      \includegraphics[width=1.0\linewidth]{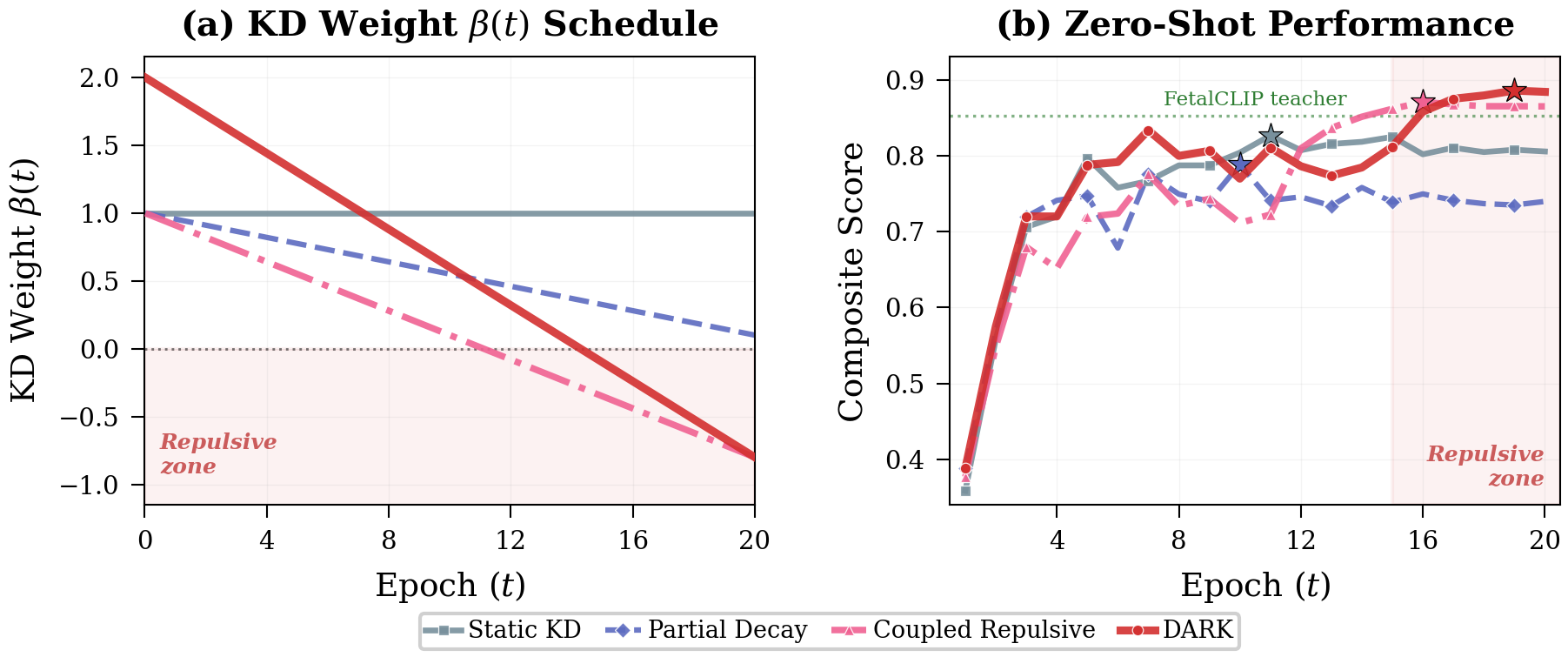}
      \caption{%
        Training dynamics for representative KD configurations. \textbf{(a)}~KD weight $\beta(t)$ schedule; repulsive variants cross below zero. \textbf{(b)}~Composite Score over epochs; repulsive runs show a late surge after the sign flip, with DARK achieving the highest final score, surpassing the teacher.
      }
      \label{fig:dynamics}
    \end{minipage}\hfill
    \begin{minipage}[t]{0.49\linewidth}
        \centering
      \includegraphics[height=2.9cm, width=1.0\linewidth]{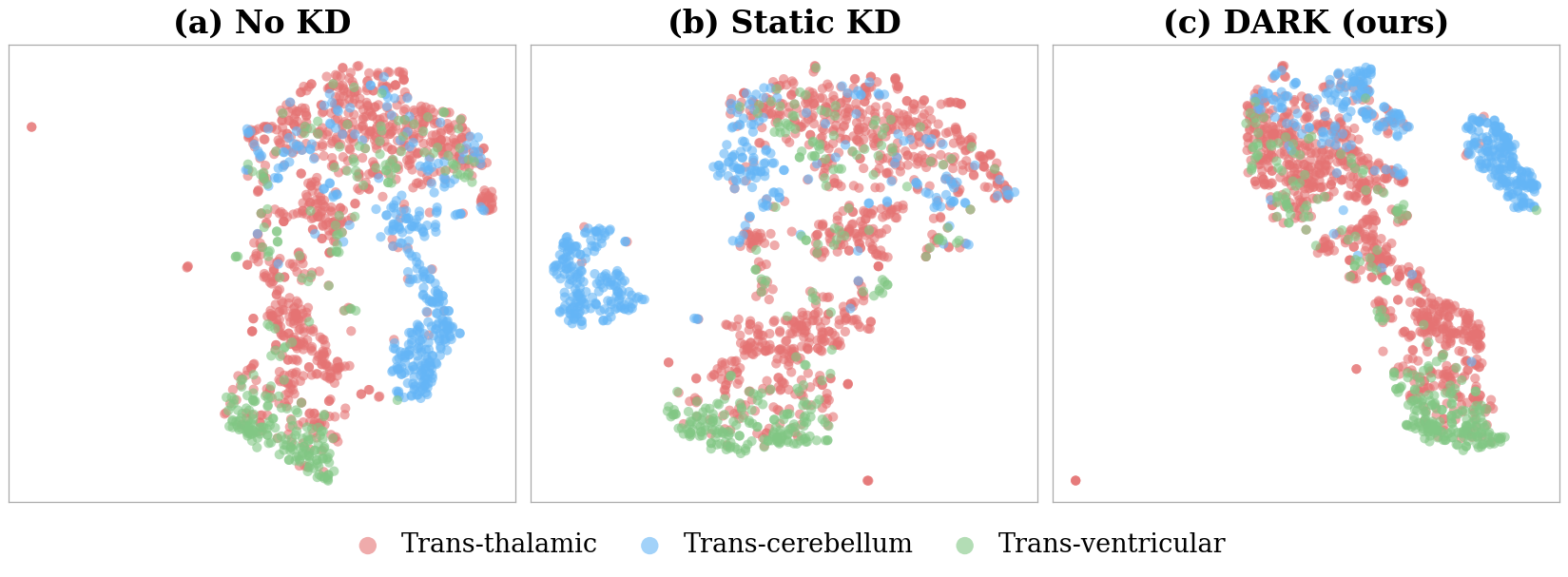}
      \caption{%
        t-SNE projections of brain sub-plane embeddings (transthalamic, transcerebellum, transventricular).
        \textbf{(a)} No KD (using CLIP loss only): overlapping clusters.
        \textbf{(b)} Static KD: marginal improvement.
        \textbf{(c)} DARK (ours): well-separated, compact clusters.
      }
      \label{fig:tsne}
    \end{minipage}
\end{figure}

\noindent\textbf{t-SNE Visualisation.}
\Cref{fig:tsne} shows t-SNE projections of brain sub-plane embeddings, the hardest subtask (three visually similar fetal head planes). Without KD, transthalamic and transventricular overlap substantially; static KD provides marginal improvement. DARK KD produces dramatically tighter, well-separated clusters, consistent with the $+$8.2\% gain in \fbrain over the teacher.

\noindent\textbf{Embedding Geometry.}
We perform quantitative cluster and spectral analysis on the Planes DB 5-plane evaluation set. \Cref{tab:geometry} reports silhouette score, inter-class cosine similarity, uniformity~\cite{wang2020uniformity}, and effective dimensionality $d_\text{eff}$ via the participation ratio of the embedding covariance spectrum. DARK KD dramatically improves cluster geometry: silhouette scores increase by $+40\%$ over Static KD (0.525 vs.\ 0.375), and inter-class cosine collapses from 0.445 to near zero. The confidence penalty provides only modest improvement in silhouette score (0.406), confirming that undirected entropy cannot match teacher-informed repulsion. Coupled Repulsion KD concentrates features into fewer dimensions ($d_\text{eff}$ 6.4 vs.\ 8.0 for static KD), while DARK KD achieves the highest $d_\text{eff}$ (10.0) and best uniformity ($-2.308$), connecting to the \cite{zbontar2021barlow} decorrelation principle. Full spectral analysis is in supplementary~\S\,\ref{app:spectral}. Per-class F1 heatmaps and confusion matrices are in supplementary~\S\,\ref{app:confusion}.

\subsection{Discussion}
\label{sec:discussion}

\noindent\textbf{When and Why Does \textsc{DARK} Work?}
\textsc{DARK} is most effective under large capacity gaps, where strict teacher mimicry underutilizes the student. In our $26\times$ setting, controlled comparison (coupled vs.\ \textsc{DARK}, both with $\beta_{\mathrm{stop}}{=}{-}0.8$) shows that diagonal protection yields $+4.2\%$ in \vrhc and $+2.1\%$ in \fbrain (\cref{sec:ablation}). The key mechanism is \emph{structured decorrelation}: the student forms well-separated representations that diverge from the teacher’s non-target structure while preserving target alignment. Evidence: (i) \textit{embedding geometry} (\cref{tab:geometry}) improves substantially (silhouette $+40\%$, near-zero inter-class cosine, highest $d_\text{eff}$, best uniformity); (ii) \textit{logit distributions} show much lower entropy (0.666$\to$0.044) with high rank correlation ($\rho{=}0.822$), indicating sharper yet consistent predictions; and (iii) \textit{per-class performance} improves, with \fbrain reaching 0.784 ($+8.2\%$ over teacher), reflecting better fine-grained discrimination.


\noindent\textbf{Why can the student surpass the teacher?}
The teacher's ViT-L/14 distributes representational capacity across all inter-class relationships through global self-attention, including confusable pairs (\eg brain sub-planes sharing similar ultrasound textures). The compact FastViT student (11.4M visual parameters) cannot afford such distributed representations; standard KD wastes capacity, forcing the student to approximate inter-class relationships it cannot faithfully reproduce. DARK KD uses the teacher's confusion patterns as a structured signal for \emph{where} to build sharper boundaries, allowing the student to exploit its convolutional-attention architecture for local discriminative cues that the teacher's global attention did not prioritize.

\noindent\textbf{When Does It Fail?}
Repulsive KD fails when the schedule is too aggressive or poorly phased. An exploratory coupled run with excessive repulsion ($\beta_{\mathrm{stop}}{=}{-}1.6$) collapses after $\sim$epoch 14, while overly strong off-diagonal amplification ($\beta_{\mathrm{start}}{=}8$) degrades \vrhc (to 0.786). Successful schedules consistently require an initial attractive phase to absorb teacher knowledge before transitioning to repulsion. Additionally, feature KD is harmful under the $26\times$ gap (\vrhc: 0.794$\to$0.759), as the teacher’s embedding space is too mismatched for effective pointwise alignment.



\noindent\textbf{Limitations and Future Work} While retrospective benchmarks establish strong zero-shot capabilities, clinical translation requires prospective validation to assess robustness against diverse ultrasound hardware and operator variability. Given \mfc's 1.6\,ms latency, our immediate focus is real-time evaluation on point-of-care ultrasound (POCUS) devices, targeting live assistive feedback in low-resource settings. Furthermore, because DARK is architecture- and domain-agnostic, future work will extend this framework to other resource-constrained clinical applications, such as echocardiography and cross-modal retrieval in general radiology.

\section{Conclusion}
\label{sec:conclusion}
We introduced \textbf{Diagonal-Anchored Repulsive Knowledge Distillation} (DARK), based on the premise that under extreme teacher–student capacity gaps, strict imitation is inefficient and often counterproductive. DARK instead disentangles what to preserve from what to discard by anchoring alignment on matched image–text pairs while progressively repelling non-target similarities that reflect the teacher’s excess capacity and biases. Applied to fetal ultrasound, DARK enables \textbf{MobileFetalCLIP} to achieve strong zero-shot performance with a 26$\times$ smaller visual encoder and 32$\times$ fewer GMACs, surpassing its teacher on HC18 validity and brain sub-plane F1 while running in 1.6,ms on an iPhone~16~Pro. Beyond efficiency gains, the student exhibits a more structured embedding geometry, retaining confidence where needed while avoiding inherited inter-class confusion. Overall, our results suggest that under tight resource constraints, learning what \emph{not} to imitate can be as important as learning what to match, positioning repulsion-based objectives like DARK as a principled direction for compressing large multimodal models.



\bibliographystyle{plainnat}
\bibliography{main}


\appendix
\input{supplemental.tex}


\newpage
\input{checklist.tex}

\end{document}

%% file: table_sota.tex
\begin{table}[!t]
  \caption{%
    Zero-shot comparison on fetal ultrasound benchmarks. Static Logit KD implements the CLIP-KD~\cite{clipkd2024} objective. Composite Score = (F1$_{\mathrm{Pl\_Br}}$ + VR$_{\mathrm{HC18}}$)/2.
  }
  \label{tab:sota}
  \centering
  \scriptsize
  \setlength{\tabcolsep}{3pt}
  \begin{tabular}{@{}l r c c c c c c@{}}
    \toprule
    \textsc{Models} & \textsc{Params.} & \textsc{F1$_{\mathrm{5Plane}}$} & \textsc{F1$_{\mathrm{3Brain}}$} & \textsc{F1$_{\mathrm{Pl\_Br}}$} & \textsc{VR$_{\mathrm{HC18}}$} & \textsc{Composite Score} \\
    \midrule
    \textcolor{deepblue!50}{\textsc{Teacher Model}} \\
    FetalCLIP (ViT-L/14) & 427M & \textbf{0.973} & 0.702 & 0.871 & 0.835 & 0.853 \\
    \midrule
    \textcolor{deepblue!50}{\textsc{General VLMs}} \\
    CLIP (ViT-L/14) & 427M & 0.308 & 0.206 & 0.270 & 0.110 & 0.190 \\
    BiomedCLIP (ViT-B/16) & 150M & 0.603 & 0.236 & 0.466 & 0.240 & 0.353 \\
    UniMed-CLIP (ViT-B/16) & 150M & 0.679 & 0.187 & 0.495 & 0.090 & 0.293 \\
    \midrule
    \textcolor{deepblue!50}{\textsc{Supervised} (\textit{non--zero-shot~setting})} \\
    SonoNet-16 & 11M & 0.827 & 0.485 & 0.699 & -- & -- \\
    \midrule
    \textcolor{deepblue!50}{\textsc{\mfc - (Student)}} \\
    Using $\mathcal{L}_{\mathrm{CLIP}}$ only (without $\mathcal{L}_{\mathrm{KD}}$) & 75M & 0.889 & 0.712 & 0.823 & 0.713 & 0.768 \\
    Static Logit KD~~ ($\mathcal{L}_{\mathrm{CLIP}}+1.0\cdot\mathcal{L}_{\mathrm{KD}}$) & 75M & 0.946 & 0.715 & 0.859 & 0.794 & 0.826 \\
    Coupled Repulsive KD~~ ($\mathcal{L}_{\mathrm{CLIP}}+\beta(t)\cdot\mathcal{L}_{\mathrm{KD}}$) & 75M & 0.933 & 0.763 & 0.869 & 0.844 & 0.857 \\
    Diagonal-Anchored Repulsive KD (DARK) & 75M & 0.946 & \textbf{0.784} & \textbf{0.886} & \textbf{0.886} & \textbf{0.886} \\
    \bottomrule
  \end{tabular}
\end{table}

%% file: table_ablation.tex
\begin{table}[!h]
  \caption{%
    Ablation on KD Strategies: KD strategies for \mfc (FastViT-MCI0, 11.4M visual encoder) distilled from FetalCLIP ViT-L/14 (304M). Composite Score = (F1$_{\mathrm{Pl\_Br}}$ + VR$_{\mathrm{HC18}}$)/2.
  }
  \label{tab:ablation}
  \centering
  \scriptsize
  \setlength{\tabcolsep}{2.5pt}
  \begin{tabular}{@{}lccccc@{}}
    \toprule
    \textsc{Configurations} & \textsc{F1$_{\mathrm{5Plane}}$} & \textsc{F1$_{\mathrm{3Brain}}$} & \textsc{F1$_{\mathrm{Pl\_Br}}$} & \textsc{VR$_{\mathrm{HC18}}$} & \textsc{Composite~Score}\\
    \midrule
    FetalCLIP (ViT-L/14) - Teacher & 0.973 & 0.702 & 0.871 & 0.835 & 0.853 \\
    \midrule
    Using $\mathcal{L}_{\mathrm{CLIP}}$ only (without $\mathcal{L}_{\mathrm{KD}}$) & 0.889 & 0.712 & 0.823 & 0.713 & 0.768 \\
    Static KD ($\mathcal{L}_{\mathrm{CLIP}}+\lambda\cdot\mathcal{L}_{\mathrm{KD}}$), $\lambda{=}1$ & 0.946 & 0.715 & 0.860 & 0.794 & 0.826 \\
    Partial KD Annealing ($\lambda:\,1{\to}0.1$) & 0.902 & 0.713 & 0.831 & 0.746 & 0.788 \\
    Full KD Annealing ($\lambda:1{\to}0$) & 0.842 & 0.742 & 0.805 & 0.731 & 0.768 \\
    Static KD + Feature KD  & 0.946 & 0.664 & 0.840 & 0.759 & 0.800 \\
    Confidence Penalty ($\varepsilon{=}0.1$) & 0.854 & 0.680 & 0.789 & 0.749 & 0.769 \\
    \midrule
    \multicolumn{6}{@{}l}{{Coupled Repulsive KD~($\mathcal{L}_{\mathrm{CLIP}}+\beta(t)\cdot\mathcal{L}_{\mathrm{KD}}$)}} \\
    \quad $(\beta_{\mathrm{start}}{=}2$,~~$\beta_{\mathrm{stop}}{=}{-}0.8)$ & 0.933 & 0.763 & 0.869 & 0.844 & 0.857 \\
    \midrule
    \multicolumn{6}{@{}l}{Diagonal-Anchored Repulsive KD~   $(\mathcal{L}_\mathrm{CLIP}\;+\;\mathcal{L}_\mathrm{KD}^{\mathrm{diag}}\;+\;\beta(t)\cdot\mathcal{L}_\mathrm{KD}^{\mathrm{off}\text{-}\mathrm{diag}})$} \\
     \quad DARK $(\beta_{\mathrm{start}}{=}2$,~~$\beta_{\mathrm{stop}}{=}{-}0.8)$ & \textbf{0.946} & \textbf{0.784} & \textbf{0.885} & \textbf{0.886} & \textbf{0.886} \\
    \midrule
    Stronger amplification ($\beta_{\mathrm{stop}}{=}-0.8$) \\
    \quad $\beta_{\mathrm{start}}{=}4$ & 0.950 & 0.709 & 0.860 & 0.854 & 0.857 \\
    \quad $\beta_{\mathrm{start}}{=}8$ & 0.943 & 0.714 & 0.857 & 0.786 & 0.822 \\
    \midrule
    Weaker repulsion ($\beta_{\mathrm{start}}{=}2$) \\
    \quad $\beta_{\mathrm{stop}}{=}{-}0.4$ & 0.938 & 0.724 & 0.858 & 0.784 & 0.821 \\
    \quad $\beta_{\mathrm{stop}}{=}{-}0.5$ & 0.941 & 0.725 & 0.860 & 0.846 & 0.853 \\
    \bottomrule
  \end{tabular}
\end{table}

%% file: supplemental.tex

\clearpage

\section*{DARK: Diagonal-Anchored Repulsive Knowledge Distillation for Vision-Language Models under Extreme Compression}

\section*{Supplementary Material}


Our supplementary material is organized as follows:

\begin{itemize}
    \setlength\itemindent{-2em}
    \setlength\leftmargin{0pt}
    \renewcommand\labelitemi{}
    \setlength\itemsep{0pt}

\item \hyperref[app:impl]{\textbf{A}~~Implementation Details}
\begin{itemize}
    \setlength\itemindent{-1em}
    \setlength\leftmargin{0pt}
    \setlength\itemsep{0pt}
    \item \hyperref[sub:model_arch]{Model Architecture}
    \item \hyperref[sub:train_config]{Training Configuration}
    \item \hyperref[app:kd_form]{KD Objective Form}
    \item \hyperref[sub:data_pipeline]{Data Pipeline}
    \item \hyperref[sub:eval_metric]{Evaluation Metrics}
\end{itemize}

\item \hyperref[app:additional]{\textbf{B}~~Additional Analysis}
\begin{itemize}
    \setlength\itemindent{-1em}
    \setlength\leftmargin{0pt}
    \setlength\itemsep{0pt}
    \item \hyperref[app:confusion]{Confusion Matrices and Per-Class Analysis}
    \item \hyperref[app:logit_dist]{Logit Distribution Analysis}
    \item \hyperref[app:dynamics_all]{Full Training Dynamics}
    \item \hyperref[app:dkd]{Relationship to Decoupled Knowledge Distillation}
    \item \hyperref[app:loss_components]{Loss Component Analysis}
    \item \hyperref[app:spectral]{Spectral Analysis of Learned Embeddings}
    \item \hyperref[app:linear_probe]{Linear Probing Evaluation}
\end{itemize}

\item \hyperref[app:eval]{\textbf{C}~~Evaluation Protocol and Seed Robustness}
\begin{itemize}
    \setlength\itemindent{-1em}
    \setlength\leftmargin{0pt}
    \setlength\itemsep{0pt}
    \item \hyperref[sub:zero_shot]{Zero-Shot Classification}
    \item \hyperref[sub:hc18_validity]{HC18 Validity Rate}
    \item \hyperref[app:seeds]{Seed Robustness}
\end{itemize}
\item \hyperref[checklist]{\textbf{D}~~NeurIPS Paper Checklist} 
\end{itemize}

\setcounter{section}{0}
\setcounter{subsection}{0}
\setcounter{figure}{0}
\setcounter{table}{0}
\renewcommand{\thesection}{\Alph{section}}
\renewcommand{\thesubsection}{\thesection.\arabic{subsection}}
\renewcommand{\thefigure}{S\arabic{figure}}
\renewcommand{\thetable}{S\arabic{table}}
\renewcommand{\theHsection}{supp.\arabic{section}}
\renewcommand{\theHsubsection}{supp.\arabic{section}.\arabic{subsection}}
\renewcommand{\theHfigure}{supp.\arabic{figure}}
\renewcommand{\theHtable}{supp.\arabic{table}}

\noindent

\section{Implementation Details}
\label{app:impl}

\subsection{Model Architecture}
\label{sub:model_arch}
\noindent\textbf{Student image encoder.}
FastViT-MCI0~\cite{fastvit2023,mobileclip2024} is a hybrid CNN-ViT architecture with structural reparameterisation.
Input resolution: $256{\times}256$; output embedding dimension: 512; 11.4M image encoder parameters ($\approx$75M total including the text encoder).

\noindent\textbf{Student text encoder.}
CLIP-style Transformer (4-layer) with embedding dimension 512, 4 attention heads, and vocabulary size 49{,}408.
Weights are initialised from the pre-trained MobileCLIP S0 checkpoint~\cite{mobileclip2024}.

\noindent\textbf{Teacher.}
FetalCLIP~\cite{fetalclip2025} with a ViT-L/14 image encoder.
Input resolution: $224{\times}224$; output embedding dimension: 768; $\approx$304M image encoder parameters ($\approx$427M total).
The teacher is loaded with \texttt{force\_context\_length=117} and remains frozen throughout training.

\subsection{Training Configuration}
\label{sub:train_config}
\Cref{tab:hyperparams} summarises the full training configuration used across all reported runs.
Experiment-specific values (starting amplification value $\beta_{\mathrm{start}}$, lowest repulsion factor $\beta_{\mathrm{stop}}$, feature KD weight $\lfeat$, confidence penalty $\varepsilon$) are varied as described in the main-paper ablation (Table~\ref{tab:ablation} therein) and are listed alongside the base configuration.

\begin{table}[tb]
  \caption{Full training hyperparameter configuration. Values in the right column apply to all runs unless noted otherwise.}
  \label{tab:hyperparams}
  \centering
  \begin{tabular}{@{}p{0.38\linewidth}p{0.56\linewidth}@{}}
    \toprule
    Hyperparameter & Value \\
    \midrule
    Learning rate ($\eta_\mathrm{peak}$) & $1 \times 10^{-5}$ \\
    LR schedule & Cosine decay with linear warmup (500 warmup steps; world-size adjusted) \\
    Weight decay & 0.1 \\
    Batch size & 512 \\
    Gradient accumulation steps & 2 \\
    Effective batch size & 1{,}024 (typical global batch) \\
    Training epochs & 20 \\
    KD temperature $\tau_\mathrm{KD}$ & 5.0 \\
    Logit KD initial weight $\lambda_0$ & 1.0 \\
    Distill-weight decay & Enabled for decay/repulsive runs; disabled for static KD baselines \\
    Maximum distillation weight $\beta_{\mathrm{start}}$ & 1.0 or 2.0 (default) / 4.0 / 8.0 (repulsive experiments) \\
    Minimum distillation weight $\beta_{\mathrm{stop}}$ & 0.1 / 0.0 / $-$0.5 / $-$0.8 / $-$1.0 (plus static runs without decay) \\
    Decoupled KD mode & Disabled (coupled, default) or diagonal-protected (selective repulsive) \\
    Feature KD weight $\lfeat$ & 0.0 (logit-only) or 2000 (MSE feature KD, CLIP-KD style) \\
    Confidence penalty $\varepsilon$ & 0.0 (default) or 0.1 (confidence-penalty ablation) \\
    Mixed precision & FP16 \\
    Gradient clipping & 1.0 (global $\ell_2$ norm) \\
    Optimiser & AdamW~\cite{loshchilov2019adamw} ($\beta_1{=}0.9$, $\beta_2{=}0.98$, $\epsilon{=}10^{-6}$) \\
    Framework & HuggingFace Accelerate~\cite{accelerate2022}, DDP \\
    Seed & 42 (default; see \S\,\ref{app:seeds} for multi-seed results) \\
    GPU & NVIDIA RTX PRO 5000 Blackwell \\
    \bottomrule
  \end{tabular}
\end{table}

\subsection{KD Objective Form}
\label{app:kd_form}

The implementation uses teacher-only temperature softening for logit KD.
Given the student's $N{\times}N$ similarity logit matrix $S^{\mathrm{S}}$ and the teacher's $S^{\mathrm{T}}$, we compute row-wise softmax distributions:
\begin{equation}
  p^{\mathrm{T}}_i = \mathrm{softmax}\!\bigl(S^{\mathrm{T}}_{i,:}/\tau_\mathrm{KD}\bigr),
  \qquad
  q^{\mathrm{S}}_i = \mathrm{softmax}\!\bigl(S^{\mathrm{S}}_{i,:}\bigr),
\end{equation}
where the teacher is softened with KD temperature $\tau_\mathrm{KD}{=}5.0$ and the student operates at its native logit scale. The symmetric distillation loss averages the image-to-text $(I\to T)$ and text-to-image $(T\to I)$ directions of the similarity matrix,
\begin{equation}
  \mathcal{L}_\mathrm{KD}
  = \tfrac{1}{2}\!\left[\,
      \mathcal{H}\!\left(p^{\mathrm{T}}, q^{\mathrm{S}}\right)_{I\to T}
      + \mathcal{H}\!\left(p^{\mathrm{T}}, q^{\mathrm{S}}\right)_{T\to I}
   \,\right],
  \label{eq:kd}
\end{equation}
where $\mathcal{H}(p,q)=-\sum_j p_j\log q_j$ is the cross-entropy, which is equivalent to KL divergence $\mathrm{KL}\!\left(p^{\mathrm{T}}\,\|\,q^{\mathrm{S}}\right)$ up to additive teacher-entropy constants.
\begin{equation}
  \mathcal{H}(p^{\mathrm{T}}, q^{\mathrm{S}}) = \mathrm{KL}(p^{\mathrm{T}} \| q^{\mathrm{S}}) + \mathcal{H}(p^{\mathrm{T}}).
\end{equation}

Since $\mathcal{H}(p^{\mathrm{T}})$ does not depend on student parameters, the optimisation gradients are identical.
Importantly, we do \emph{not} apply student-side temperature scaling and therefore do not use the classical $T^2$ prefactor from Hinton-style KD~\cite{hinton2015kd}.
The rationale is that in the contrastive $N{\times}N$ setting, the student's learned logit scale (inverse temperature) is itself a trainable parameter and already calibrates the sharpness of the student's similarity distribution.
Applying an additional $\tau_\mathrm{KD}$ to the student would interfere with this learned calibration.

\subsection{Data Pipeline}
\label{sub:data_pipeline}

\paragraph{Training data curation.}
We curated a total of 246{,}349 fetal ultrasound image-caption pairs from two complementary sources. The primary source contains 225{,}429 fetal ultrasound images collected from a tertiary hospital, spanning 6{,}493 patients with a mean gestational age of 148 $\pm$ 16 days. All of the data from this source contain information about gestational age and pixel spacing. 161{,}033 of them contain diverse anatomical labels from clinicians, and 64{,}396 of them contain anatomical pseudo-labels. To generate captions for each image, we used GPT-4o to generate templates for every anatomical label, and each template contains placeholders for gestational age and pixel spacing, such as \textit{"Fetal ultrasound image at \{weeks\} weeks and \{days\} days gestational age, focusing on the abdomen with a pixel spacing of \{pixel\_spacing\} mm/pixel"} for images with an abdomen label. In addition to this source, we add 2{,}092 image-caption pairs derived from a fetal ultrasound textbook and upsample them 10 times during training to increase their sampling frequency. Due to regulation, privacy constraints, and copyright restrictions from the textbook author, the training dataset is not publicly released.

Training data is served in WebDataset~\cite{webdataset2021} format (25 tar shards, 246{,}349 image-caption pairs total).
The augmentation pipeline is:
\begin{enumerate}
  \item Random affine (rotation $\pm7^\circ$, translation 0.05 in each axis, scale fixed at 1.0)
  \item Random colour jitter (brightness 0.15, contrast 0.15, saturation 0.15)
  \item Student branch: \texttt{Resize(256)}$\to$\texttt{CenterCrop(256)}
  \item Teacher branch (KD mode): \texttt{Resize(224)} without centre crop
  \item Normalisation: student uses ImageNet statistics; teacher uses OpenAI CLIP statistics
\end{enumerate}
In KD mode, both student and teacher views are generated from the same random seed per image (\emph{coupled augmentation}), ensuring the $N{\times}N$ logit matrices compare semantically equivalent views.

\subsection{Evaluation Metrics}
\label{sub:eval_metric}
\paragraph{F1-score.}
We utilize the F1-score as the primary metric for the classification tasks. We compute precision, recall, and F1-score as follows:
\begin{equation}
    \mathrm{Precision} = \frac{\mathrm{TP}}{\mathrm{TP} + \mathrm{FP}},
    \qquad
    \mathrm{Recall} = \frac{\mathrm{TP}}{\mathrm{TP} + \mathrm{FN}},
\end{equation}
\begin{equation}
    \mathrm{F1} =
    \frac{2 \times \mathrm{Precision} \times \mathrm{Recall}}
    {\mathrm{Precision} + \mathrm{Recall}},
\end{equation}
where $\mathrm{TP}$, $\mathrm{FP}$, and $\mathrm{FN}$ denote the number of true positives, false positives, and false negatives, respectively.

\paragraph{Macro F1-score.}
Our classification tasks involve multi-class classification, i.e. five fetal plane classes
($\mathcal{C}_{\mathrm{5Plane}} = \{\mathrm{abdomen}, \mathrm{brain}, \mathrm{femur}, \mathrm{thorax},  \mathrm{cervix} \}$)
and three fetal brain sub-plane classes ($\mathcal{C}_{\mathrm{3Brain}} = \{\mathrm{transthalamic}, \mathrm{transcerebellum}, \mathrm{transventricular} \}$).
To aggregate the F1-scores for each classification task, we compute the F1-score for each class and then average the class-wise F1-scores as follows:
\begin{equation}
    \mathrm{F1}_\mathrm{5Plane}
    =
    \frac{1}{|\mathcal{C}_{\mathrm{5Plane}}|}
    \sum_{c \in \mathcal{C}_{\mathrm{5Plane}}} \mathrm{F1}_c,
    \qquad
    \mathrm{F1}_\mathrm{3Brain}
    =
    \frac{1}{|\mathcal{C}_{\mathrm{3Brain}}|}
    \sum_{c \in \mathcal{C}_{\mathrm{3Brain}}} \mathrm{F1}_c.
\end{equation}
This macro averaging assigns equal importance for each class.

\paragraph{Combined Fetal Plane F1-score.}
We additionally report an aggregate fetal plane F1-score, denoted by $\mathrm{F1}_\mathrm{Pl\_Br}$, which combines $\mathrm{F1}_\mathrm{5Plane}$ and $\mathrm{F1}_\mathrm{3Brain}$. $\mathrm{F1}_\mathrm{Pl\_Br}$ is computed by taking a weighted average of $\mathrm{F1}_\mathrm{5Plane}$ and $\mathrm{F1}_\mathrm{3Brain}$ as follows:
\begin{equation}
    \mathrm{F1}_\mathrm{Pl\_Br}
    =
    \frac{5 \times \mathrm{F1}_\mathrm{5Plane}
    +
    3 \times \mathrm{F1}_\mathrm{3Brain}}{8}.
\end{equation}
This corresponds to the macro F1-score across all eight classes.

\paragraph{HC18 Validity Rate.}
We follow the FetalCLIP (\cite{fetalclip2025}) procedure to report the HC18 validity rate ($\mathrm{VR}_{\mathrm{HC18}}$). To predict gestational age (GA) from a fetal ultrasound image, we first extract embeddings from the image and from text prompts that include gestational age information ranging from 14 to 40 weeks. The predicted gestational age is the one which corresponds to the text prompt with the highest similarity to the image in the embedding space. Since the ground-truth gestational age is not available, we use a proxy to assess whether the prediction is valid based on the available ground-truth head circumference. The relation between head circumference and gestational age can be represented as (\cite{fetalcalculator}):
\begin{equation}
\label{eq:hc_ga_relationship}
\mathrm{HC} = b_0 + b_1t + b_2t^2 + b_3t^3 + b_4t^4,
\end{equation}
where $t$ denotes gestational age in days and $b_{(\cdot)}$ are quantile regression constants. Using Eq. \eqref{eq:hc_ga_relationship}, we obtain the 2.5th and 97.5th percentile head circumference bounds for the predicted gestational age. A gestational age prediction is considered valid if the ground-truth head circumference falls within these predicted head circumference bounds. The $\mathrm{VR}_{\mathrm{HC18}}$ is then computed as
\begin{equation}
    \mathrm{VR}_{\mathrm{HC18}}
    =
    \frac{1}{N}
    \sum_{i=1}^{N}
    \mathbb{I}
    \left[
    \mathrm{HC}_i
    \in
    \left(
    P_{2.5}(\widehat{\mathrm{GA}}_i),
    P_{97.5}(\widehat{\mathrm{GA}}_i)
    \right)
    \right],
\end{equation}
where $N$ is the number of evaluated samples, $\mathrm{HC}_i$ is the ground-truth head circumference for image $i$, $\widehat{\mathrm{GA}}_i$ is the predicted gestational age, $P_{2.5}(\widehat{\mathrm{GA}}_i)$ and $P_{97.5}(\widehat{\mathrm{GA}}_i)$ denote the 2.5th and 97.5th percentile head circumference bounds computed from the predicted gestational age using Eq. \eqref{eq:hc_ga_relationship}, and $\mathbb{I}[\cdot]$ is the indicator function.

\paragraph{Composite Score.}
We utilize the average of \fplanebrain and the \vrhc as overall performance score:
\begin{equation}
    \text{Composite Score}
    =
    \frac{
    \mathrm{F1}_\mathrm{Pl\_Br}
    +
    \mathrm{VR}_{\mathrm{HC18}}
    }{2}.
\end{equation}

\section{Additional Analysis}
\label{app:additional}

\subsection{Confusion Matrices and Per-Class Analysis}
\label{app:confusion}

\begin{figure}[!t]
  \centering
  \includegraphics[width=0.94\linewidth]{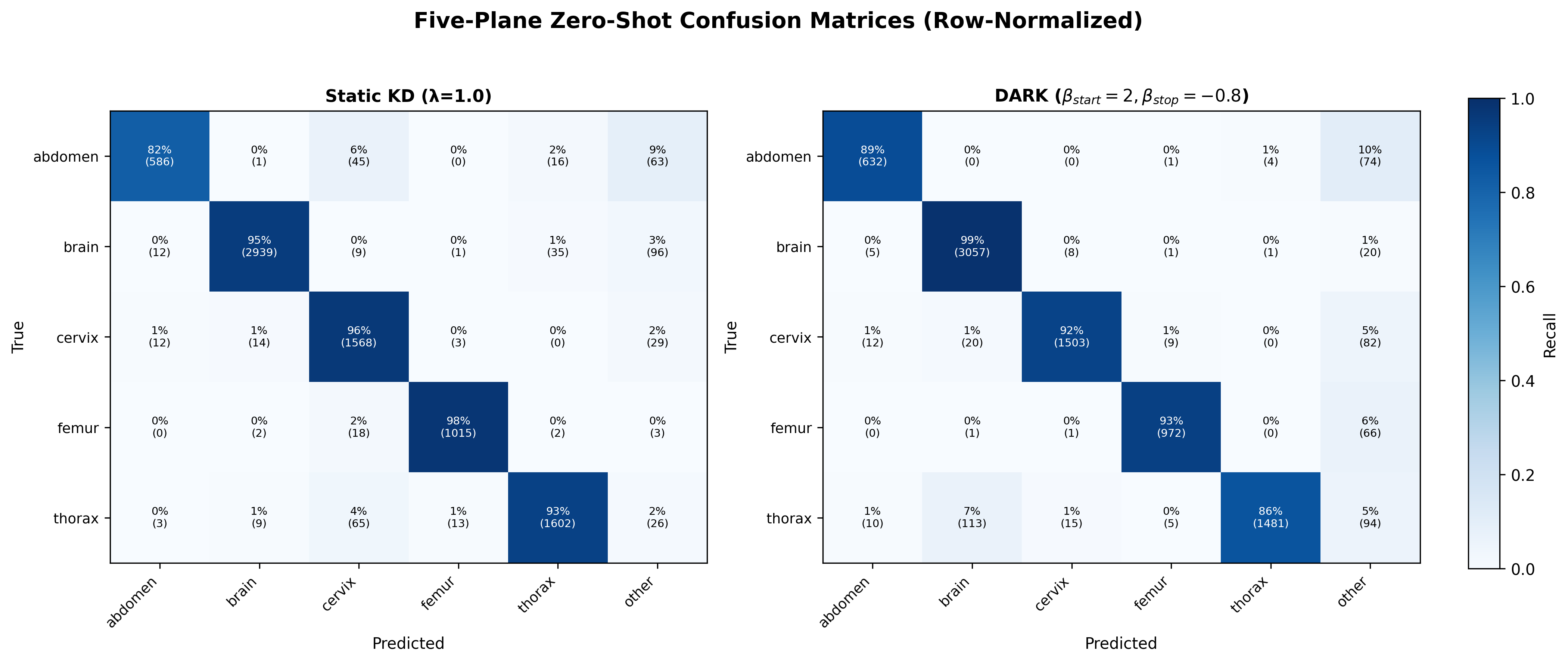}
  \caption{%
    Row-normalized confusion matrices for zero-shot 5-plane classification, with an additional ``other'' column capturing predictions to off-set classes (kidney, lips/nose, profile, spine).
    \textbf{Left:} Static KD ($\lambda{=}1.0$) achieves high recall for cervix (96\%), brain (95\%), femur (98\%), and thorax (93\%), but lower abdomen recall (82\%) and 9\% off-class leakage.
    \textbf{Right:} \textsc{DARK} ($\beta_{\mathrm{start}}{=}2$, $\beta_{\mathrm{stop}}{=}{-}0.8$) improves brain (99\%) and abdomen (89\%) with minor trade-offs on cervix, femur, and thorax; the main shift is a thorax$\to$brain redistribution (1\%$\to$7\%).
  }
  \label{fig:confusion}
\end{figure}

\Cref{fig:confusion} shows row-normalised confusion matrices for zero-shot 5-plane classification, comparing Static KD ($\lambda{=}1.0$) against the best DARK configuration ($\beta_{\mathrm{start}}{=}2$, $\beta_{\mathrm{stop}}{=}{-}0.8$).

Static KD already achieves high per-class recall on the dominant planes (cervix 96\%, brain 95\%, femur 98\%, thorax 93\%) but underperforms on abdomen (82\% recall, with 9\% of abdomen samples falling outside the five-plane label set). DARK redistributes capacity rather than uniformly improving recall: brain recall rises to 99\% (vs.\ 95\%) and abdomen recall to 89\% (vs.\ 82\%), at the cost of a small drop on the remaining classes (cervix $96{\to}92$\%, femur $98{\to}93$\%, thorax $93{\to}86$\%). The most prominent local trade-off is thorax$\to$brain leakage, which grows from 1\% (9 samples, Static) to 7\% (113 samples, Selective). This pattern is consistent with the repulsive mechanism effect under a 26$\times$ capacity gap: the inverted off-diagonal gradient frees the student to sharpen brain and abdomen boundaries that the teacher's globally-averaged attention does not prioritize, at the price of slightly blurring the acoustically similar thorax--brain pair. In clinical terms, the thorax$\to$brain leak is mild --- both are standard axial planes and seldom diagnostically catastrophic when interchanged --- whereas the gain on brain recall directly supports the brain sub-plane improvements reported in the main paper. The macro-averaged 5-plane F1 (\fplane) is essentially identical for the two methods (0.945 Static KD vs.\ 0.946 DARK), confirming that DARK reshapes the per-class error pattern rather than uniformly raising or lowering accuracy; the headline contribution is the brain sub-plane gain ($+8.2\%$ over the teacher) and the \vrhc improvement, both of which are downstream of this redistribution.

\begin{figure}[!t]
  \centering
  \includegraphics[width=0.88\linewidth]{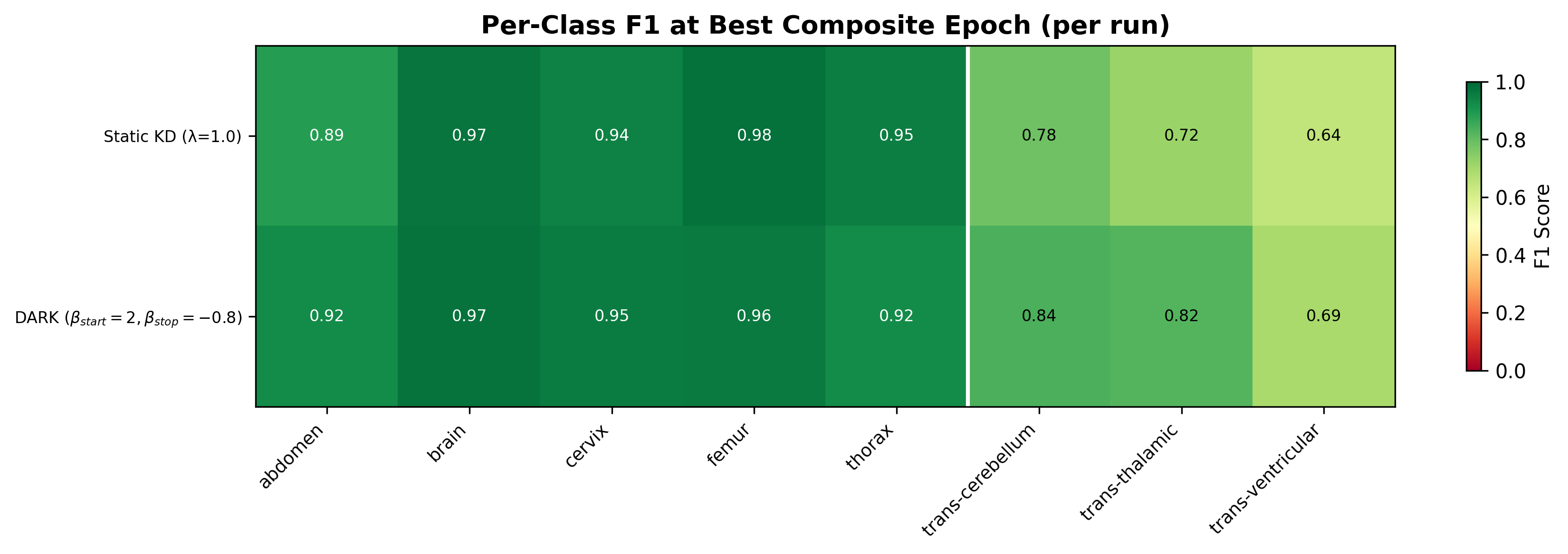}
  \caption{%
    Per-class F1 heatmap across fetal plane classes (abdomen, brain, femur, thorax, cervix) and brain sub-planes (transthalamic, transcerebellum, transventricular). Each row corresponds to the run's best epoch (selected by Composite Score). DARK ($\beta_0{=}2$, $\beta_{\mathrm{stop}}{=}{-}0.8$) achieves the strongest brain sub-plane discrimination.
  }
  \label{fig:heatmap}
\end{figure}

\Cref{fig:heatmap} shows absolute per-class F1 at each run's best epoch. DARK substantially improves \fbrain to 0.784 ($+6.9\%$ over the Static KD baseline, $+8.2\%$ over the FetalCLIP teacher), indicating that diagonal protection combined with moderate $\mathcal{L}_{\mathrm{KD}}^{\mathrm{off}\text{-}\mathrm{diag}}$ amplification is essential for fine-grained anatomical discrimination among visually similar brain cross-sections.

\subsection{Logit Distribution Analysis}
\label{app:logit_dist}

\begin{figure}[!t]
  \centering
  \includegraphics[width=0.85\linewidth]{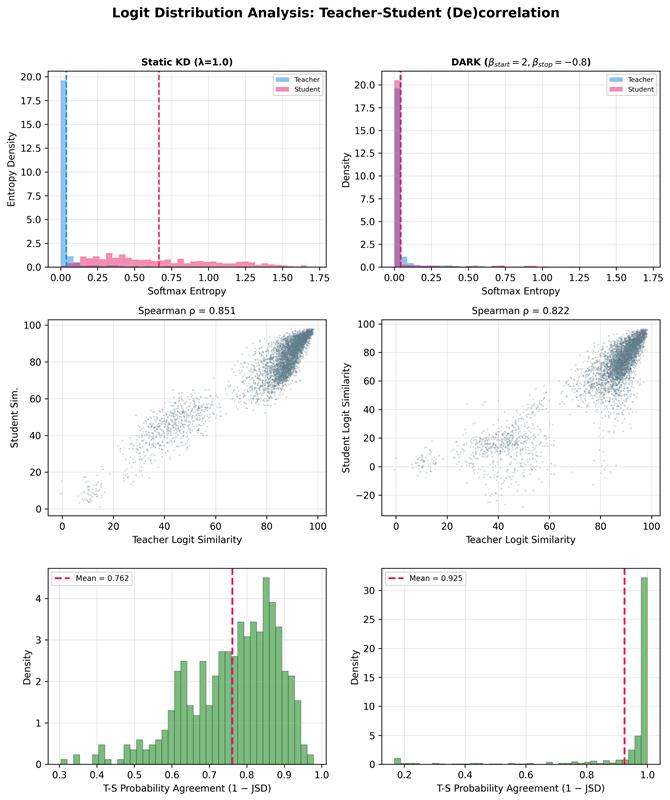}
  \caption{%
    Logit distribution analysis: teacher--student decorrelation comparing Static KD and DARK.
    \textbf{Row 1}: Softmax entropy distributions (lower = more confident; dashed lines denote empirical means).
    \textbf{Row 2}: $N{\times}N$ logit matrix rank correlation.
    \textbf{Row 3}: Per-image teacher-student probability agreement ($1 - \mathrm{JSD}$; higher = the student's per-sample class distribution matches the teacher's more closely; dashed lines denote empirical means).
  }
  \label{fig:logit_dist}
\end{figure}

\Cref{fig:logit_dist} presents the teacher--student decorrelation analysis. Three complementary metrics are computed on the first 500 samples of Planes DB (5-plane set, ``Other'' excluded) for Static KD ($\lambda{=}1.0$) and DARK ($\beta_{\mathrm{start}}{=}2$, $\beta_{\mathrm{stop}}{=}{-}0.8$). For Rows 1 and 3, the dashed vertical lines in the figure mark the empirical means computed on this same 500-sample evaluation set; the reported numbers below are taken directly from those plotted statistics.

\noindent\textbf{Row 1: Softmax entropy distributions.}
Static KD produces high student entropy (mean 0.666), indicating diffuse, uncertain predictions that spread probability mass across multiple classes.
DARK achieves a mean entropy of 0.044, indicating confident predictions concentrated on the correct class.

\noindent\textbf{Row 2: $N{\times}N$ logit matrix rank correlation.}
We compute Spearman's rank correlation $\rho$ between flattened teacher and student similarity matrices.
Static KD achieves $\rho{=}0.851$, indicating high rank-order agreement with the teacher. DARK yields $\rho{=}0.822$, still above 0.82, so the broad relational structure (which pairs are more/less similar) is preserved even under repulsion; the student diverges in \emph{specific} inter-class decisions, not in the global ranking of similarities.

\noindent\textbf{Row 3: Per-image probability agreement.}
We compute the Jensen--Shannon divergence (JSD) between the teacher's and student's 5-class probability vectors, and report agreement as $1 - \mathrm{JSD}$.
Static KD achieves mean agreement of 0.762, with predictions broadly distributed. DARK reaches 0.925, strongly concentrated near 1.0: per-image classification probabilities track the teacher's even more closely than static KD does. Combined with the lower inter-image rank correlation in Row 2 ($\rho{=}0.822$ vs.\ $0.851$), this is consistent with the design of DARK: the diagonal (matched pairs $i{=}j$) is preserved with a fixed positive weight, so the student inherits the teacher's per-sample classification confidence; the off-diagonal (non-target $i{\ne}j$) is repelled, so the inter-image similarity matrix decorrelates from the teacher's. The decorrelation is thus structured and targeted — at the level of pairwise non-target relationships, not individual class predictions.

\subsection{Full Training Dynamics}
\label{app:dynamics_all}

\begin{figure}[!t]
  \centering
  \includegraphics[width=0.92\linewidth]{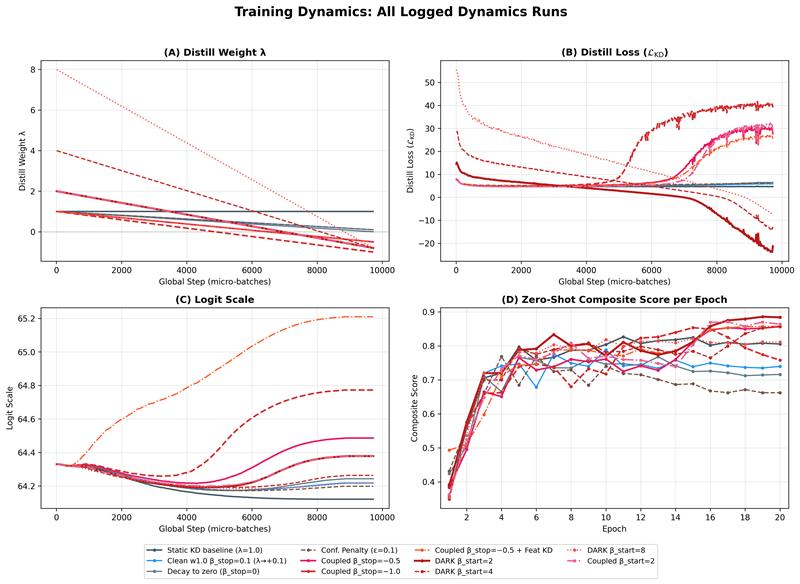}
  \caption{%
    Full training dynamics across all logged runs (4-panel overlay).
    Panels show: KD weight schedule, distill loss, learned logit scale, and the  Composite Score metric over training epochs.
    DARK ($\beta_{\mathrm{start}}{=}2$, $\beta_{\mathrm{stop}}{=}{-}0.8$) achieves the highest Composite Score, consistent with the main-paper ablation.
  }
  \label{fig:dynamics_all}
\end{figure}

\Cref{fig:dynamics_all} reports the full 4-panel training-dynamics overlay across all logged runs, including positive-decay runs, coupled repulsive variants at different repulsion values. This complements the compact 2-panel view in the main paper (\cref{fig:dynamics} therein) and confirms the same aggregate pattern: repulsive runs exhibit a characteristic late surge in zero-shot performance once the KD weight crosses zero, and DARK ($\beta_{\mathrm{start}}{=}2$, $\beta_{\mathrm{stop}}{=}{-}0.8$) achieves the highest final Composite Score among the highlighted configurations.
The auxiliary panels also show that the zero-crossing coincides with a rise in distill-loss magnitude and, for the strongest repulsive runs, a late increase in learned logit scale.

\begin{figure}[!t]
  \centering
  \includegraphics[width=0.78\linewidth]{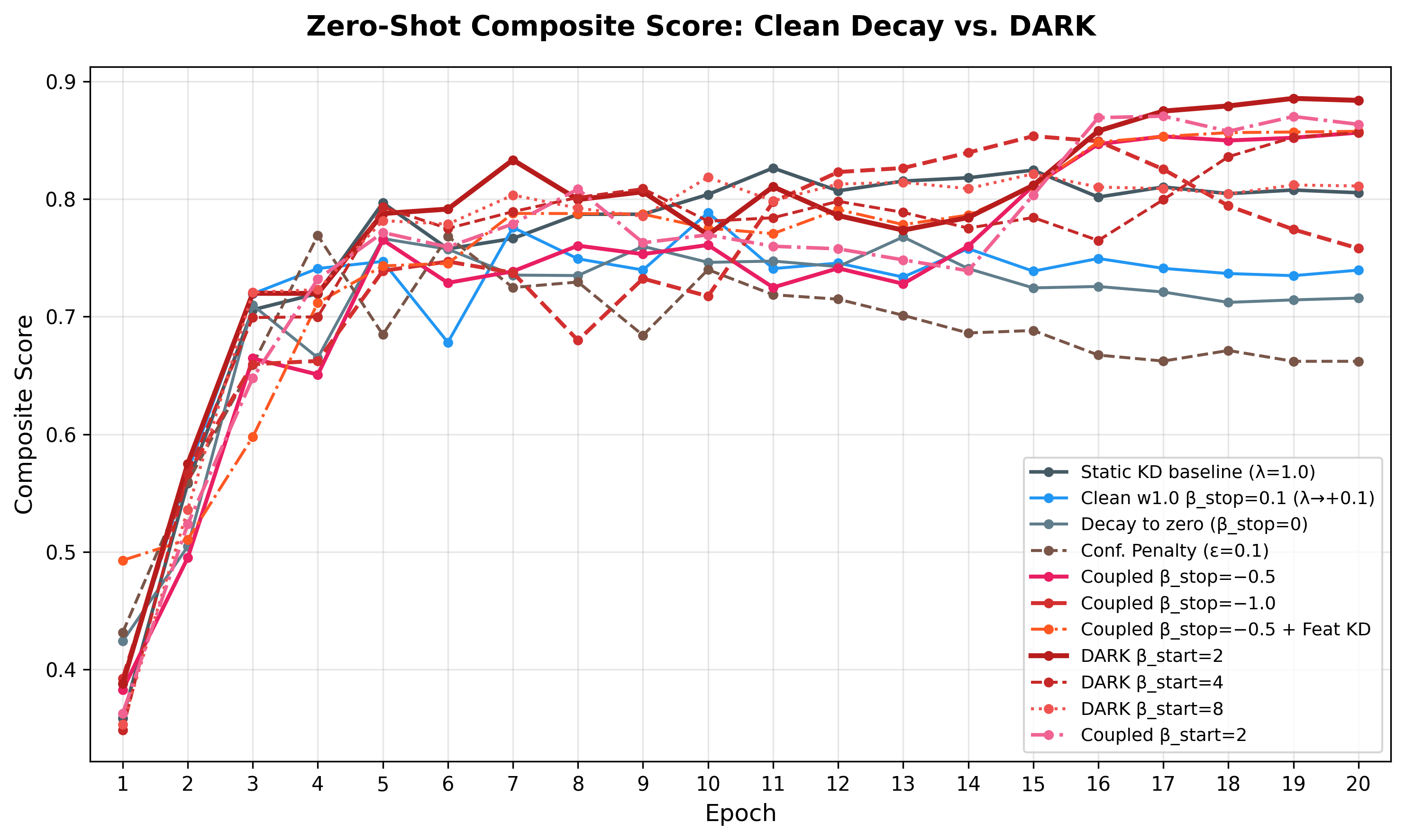}
  \caption{%
    Per-run Composite Score trajectories across training epochs.
    Static KD plateaus early, while repulsive variants continue improving through the repulsive phase.
    DARK ($\beta_{\mathrm{start}}{=}2$, $\beta_{\mathrm{stop}}{=}{-}0.8$) achieves the highest final composite score among the stable runs.
  }
  \label{fig:composite_trajectories}
\end{figure}

\Cref{fig:composite_trajectories} shows the per-run Composite Score trajectories for all configurations, providing a direct comparison of convergence behaviour. Static KD plateaus early (${\sim}$epoch 8), while repulsive variants continue improving through the repulsive phase.
DARK ($\beta_{\mathrm{start}}{=}2$, $\beta_{\mathrm{stop}}{=}{-}0.8$) exhibits the strongest late-phase acceleration, consistent with the mechanism described in the main paper: the sign flip in $\beta(t)$ converts the off-diagonal KD objective from attraction to repulsion, pushing the student to find more discriminative features.

\subsection{Relationship to Decoupled Knowledge Distillation}
\label{app:dkd}

DARK is inspired by Decoupled KD (DKD)~\cite{zhao2022dkd} but differs in several important respects due to the contrastive ($N{\times}N$) setting.
\Cref{tab:dkd_comparison} summarises the key differences.

\begin{table}[!t]
  \caption{Comparison of DARK with Zhao \etal's DKD~\cite{zhao2022dkd}.
  Both methods decompose the KD loss into target and non-target components, but differ in task formulation, weighting scheme, and, critically, the sign of the non-target weight.}
  \label{tab:dkd_comparison}
  \centering
  \begin{tabular}{@{}p{0.25\linewidth}p{0.33\linewidth}p{0.33\linewidth}@{}}
    \toprule
    Aspect & DKD~\cite{zhao2022dkd} & Ours \\
    \midrule
    Task & $C$-class classification & $N{\times}N$ contrastive similarity \\
    TCKD (diagonal) & Binary KL on $[p_t, 1{-}p_t]$ & Cross-entropy on diagonal entries ($i{=}j$) \\
    NCKD (off-diagonal) & KL on renormalised non-target distribution & Cross-entropy on off-diagonal entries ($i{\neq}j$) \\
    Weights & $\alpha{=}1$, $\beta{=}8$ (both positive, static) & Diagonal fixed at 1.0; $\beta(t)$ follows a linear schedule that decays into \emph{negative} values \\
    Temperature & Shared $\tau$ on both teacher \& student, with $T^2$ correction & Teacher-only $\tau_\mathrm{KD}$; student unscaled (see \S\,\ref{app:kd_form}) \\
    Key contribution & Independent weighting of TCKD/NCKD & Time-varying schedule with negative $\beta$ for structured repulsion \\
    \bottomrule
  \end{tabular}
\end{table}

\noindent\textbf{Critical distinction.}
DKD operates in a classification setting where both the TCKD weight $\alpha$ and the NCKD weight $\beta$ are always positive: the non-target component is upweighted to amplify ``dark knowledge'' but never inverted.
Our method extends the off-diagonal weight into \emph{negative} territory, a regime DKD does not explore.
When $\beta(t) < 0$, the gradient of the off-diagonal KD loss \emph{inverts}: instead of minimising the divergence between student and teacher non-target distributions, the objective maximises it, actively pushing the student's non-target similarity structure away from the teacher's.
Meanwhile, the fixed diagonal weight of 1.0 ensures that matched-pair alignment (the contrastive analogue of TCKD) remains stable throughout training.
This combination of protected diagonal and repulsive off-diagonal is what we term DARK.

\subsection{Loss Component Analysis}
\label{app:loss_components}

\begin{figure}[!t]
  \centering
  \includegraphics[width=0.92\linewidth]{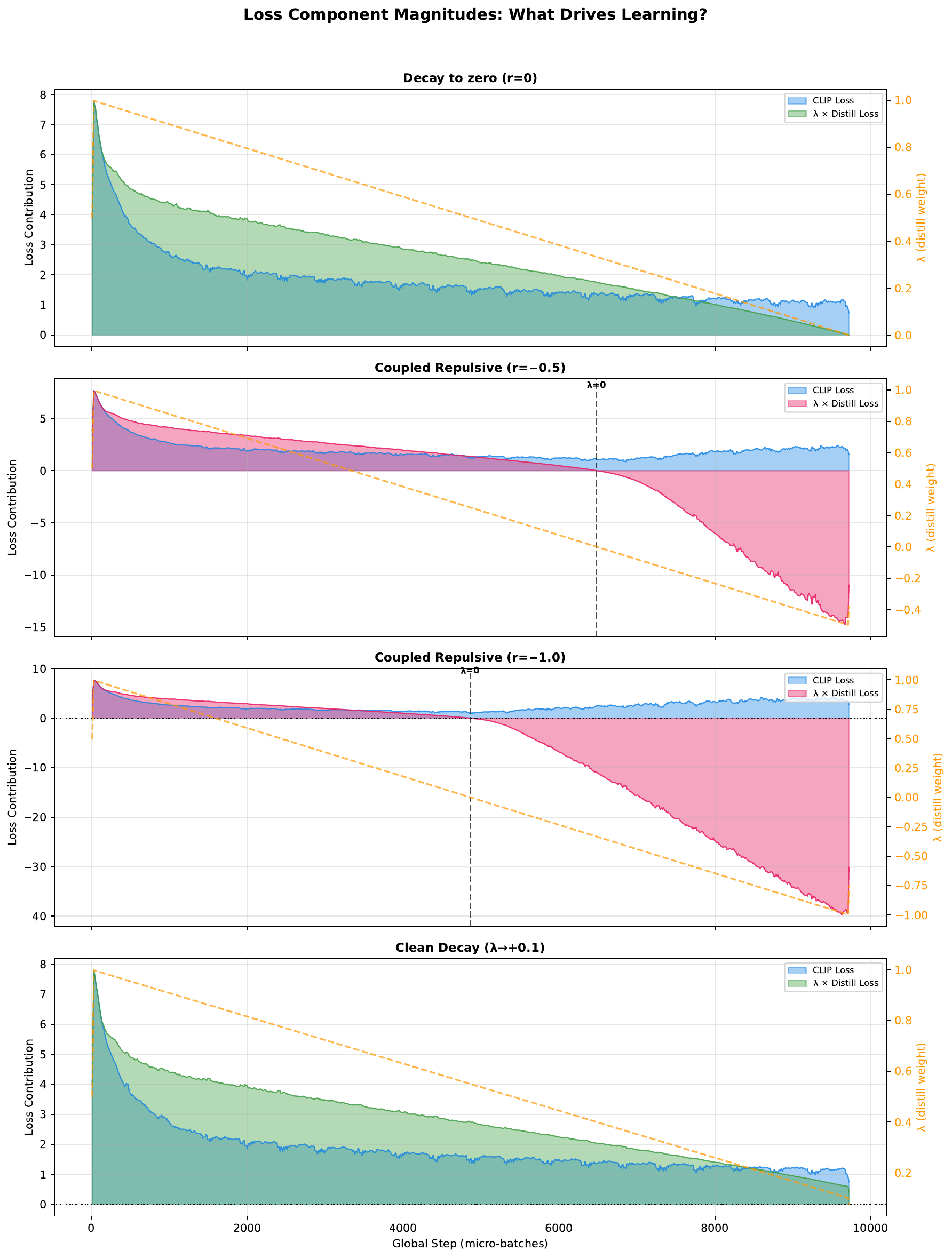}
  \caption{%
    Loss component analysis for representative decay and repulsive schedules.
    When the scheduled KD weight crosses zero (vertical dashed line), the effective KD contribution inverts.
    Moderate repulsion remains well behaved, whereas stronger repulsion produces substantially larger negative KD magnitude.
    The comparison clarifies the operating range in which repulsive KD remains stable.
  }
  \label{fig:loss_components}
\end{figure}

\Cref{fig:loss_components} compares the effective contribution of each loss component for four representative schedules: decay to zero ($\beta_{\mathrm{stop}}{=}0$), coupled repulsive KD with $\beta_{\mathrm{stop}}{=}{-}0.5$, coupled repulsive KD with $\beta_{\mathrm{stop}}{=}{-}1.0$, and a mild positive-decay run ($\beta_{\mathrm{stop}}{=}0.1$). Before the zero-crossing, both $\mathcal{L}_\mathrm{CLIP}$ and the weighted KD term contribute positively in all settings. After the zero-crossing, moderate repulsion ($\beta_{\mathrm{stop}}{=}{-}0.5$) yields a negative KD contribution while the CLIP term remains well-behaved, whereas stronger repulsion ($\beta_{\mathrm{stop}}{=}{-}1.0$) drives the weighted KD magnitude much more aggressively. These plots illustrate why mild-to-moderate repulsion is trainable in our setup, while stronger repulsive schedules require more care.

\subsection{Spectral Analysis of Learned Embeddings}
\label{app:spectral}

\begin{figure}[!t]
  \centering
  \includegraphics[width=\linewidth]{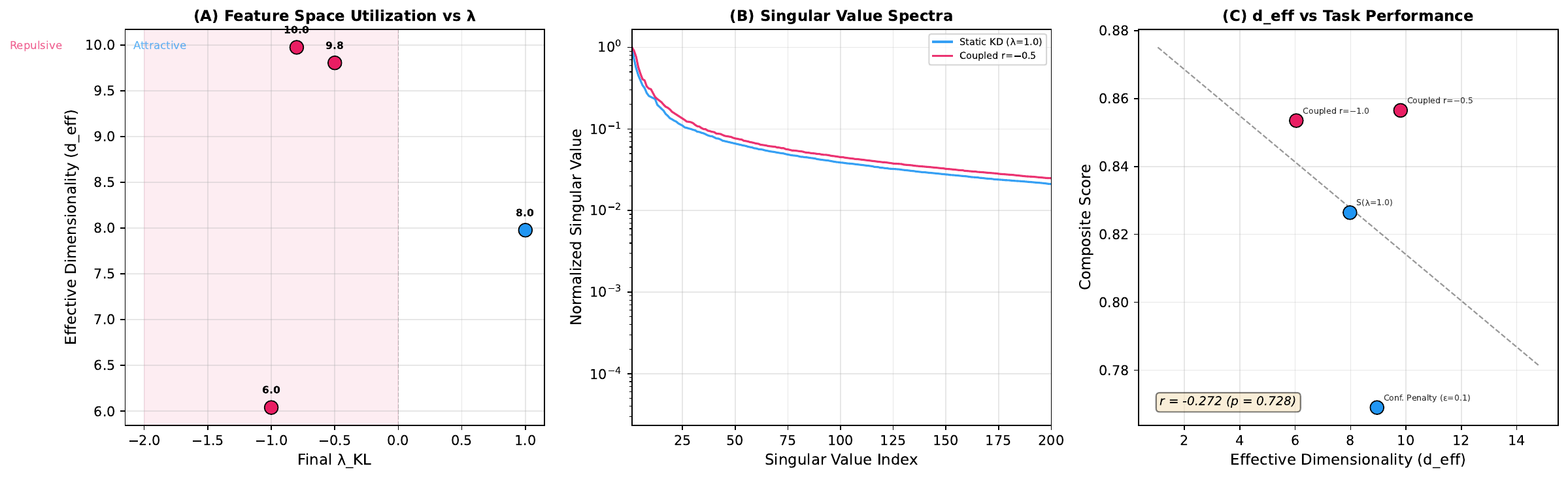}
  \caption{%
    Spectral analysis of learned embeddings (Planes DB, 5-plane, 8{,}187 images).
    \textbf{(A)}~Effective dimensionality $d_\text{eff}$ (participation ratio) across representative KD methods.
    \textbf{(B)}~Singular value spectra for Static KD and a moderate repulsive run.
    \textbf{(C)}~$d_\text{eff}$ plotted against composite task performance, illustrating that spectral richness alone does not fully determine zero-shot quality.
  }
  \label{fig:spectral}
\end{figure}

\Cref{fig:spectral} presents spectral analysis of learned embeddings, complementing the quantitative embedding geometry metrics in the main paper (Table~4 therein).
We compute the singular value decomposition of the $8{,}187 \times 512$ embedding matrix (Planes DB, 5-plane set) and derive the participation ratio as a measure of effective dimensionality $d_\text{eff}$.

The figure focuses on representative regimes rather than the full ablation.
Moderate repulsive schedules occupy the high-$d_\text{eff}$ region under negative final $\lambda$, whereas stronger repulsion ($\beta_{\mathrm{stop}}{=}{-}1.0$) drops to a much lower effective dimensionality ($d_\text{eff}{=}6.0$).
Static KD remains at $d_\text{eff}{=}8.0$, while the confidence-penalty baseline reaches higher dimensionality without matching the stronger composite scores of the repulsive runs.
The singular-value spectra in panel~B show a flatter spectrum for moderate repulsion than for static KD, consistent with broader feature-space utilisation.
Panel~C indicates that dimensionality alone does not determine downstream quality over this small set, so the spectral view should be read as complementary to the main-paper geometry metrics rather than as a standalone explanation.

\subsection{Linear Probing Evaluation}
\label{app:linear_probe}

To assess frozen feature quality independently of the contrastive alignment used in zero-shot evaluation, we conduct linear probing on three downstream tasks (\cref{tab:linear_probe}).
For each task, we freeze the image encoder, extract L2-normalized features, and train a single linear layer.

\noindent\textbf{Setup.}
\textit{6-view plane classification} and \textit{3-class brain sub-plane classification} use Planes DB~\cite{fetalplanesdb2020} with the original patient-level train/test split (7{,}129/5{,}271 for 6-view; 1{,}543/1{,}406 for brain).
\textit{Congenital heart disease (CHD) detection} uses 418 four-chamber fetal heart ultrasound videos (161 normal, 257 abnormal) with patient-stratified split (333/85); for each video, 16 frames are sampled, frame-wise features extracted, and concatenated before classification.
All experiments use 5-fold cross-validation with 5 seeds (25 runs); 95\% confidence intervals are computed via Student's $t$-distribution.

\noindent\textbf{Results.}
\mfc retains 97--98\% of the FetalCLIP teacher's linear probing performance across all three tasks, while substantially outperforming all general-purpose VLMs ($+$6--22~pp on 6-view, $+$17--22~pp on brain sub-plane, $+$5--13~pp on CHD).
The gap is expected: linear probing measures frozen feature information content, bounded by encoder capacity~\cite{stanton2021does}, whereas zero-shot evaluation probes image-text alignment, which logit-based distillation transfers most effectively~\cite{hinton2015kd,park2019rkd}.
This explains why \mfc surpasses the teacher on zero-shot tasks while retaining near-teacher linear probing.

\input{table_linear_probe}

\section{Evaluation Protocol and Seed Robustness}
\label{app:eval}

\subsection{Zero-Shot Classification}
\label{sub:zero_shot}
For each plane class, we construct a prompt ensemble following FetalCLIP's evaluation protocol~\cite{fetalclip2025}.
For a given image, we compute the cosine similarity between the image embedding and the mean text embedding for each class prompt set; the image is assigned to the class with the highest similarity.
Images are padded to square (zero-padding on the shorter axis) before resizing, matching the FetalCLIP protocol.
As in FetalCLIP, we evaluate on the full labelled pool (\texttt{split="all"}) rather than introducing an additional train/test partition.

\noindent\textbf{5-plane classification.}
8{,}187 images from Planes DB~\cite{fetalplanesdb2020} across five fetal anatomy categories: abdomen, brain, femur, thorax, and cervix (the ``Other'' category is excluded).
We report macro-average F1 (\fplane).

\noindent\textbf{Brain sub-plane classification.}
2{,}949 brain images from Planes DB across three sub-plane categories: transthalamic, transcerebellum, and transventricular.
We report macro-average F1 (\fbrain).

\subsection{HC18 Validity Rate}
\label{sub:hc18_validity}
We follow the HC18 evaluation protocol established by FetalCLIP~\cite{fetalclip2025}.
From the original 999 HC18 images~\cite{hc18dataset2018}, we retain 814 with physiologically plausible head circumference (100--342~mm, corresponding to $\approx$14--40 weeks gestational age).
Validity is computed as follows:
\begin{enumerate}
  \item For each evaluation image, compute cosine similarity between the image embedding and text embeddings for gestational age (GA)-specific prompts spanning the 14--40 week range.
  \item Assign each image to the GA week with the highest similarity score.
  \item A prediction is \emph{valid} if the ground-truth head circumference falls within the 2.5th--97.5th percentile range of the WHO fetal growth charts~\cite{kiserud2017who} for the predicted gestational age.
  \item Report the validity rate as the fraction of images with valid predictions.
\end{enumerate}
This metric jointly tests the model's ability to associate fetal head images with the correct gestational age \emph{and} the alignment between visual features and GA-specific text representations.
HC18 validity is particularly sensitive to knowledge distillation strategy because it depends on fine-grained image--text retrieval rather than coarse classification.

\subsection{Seed Robustness}
\label{app:seeds}

All ablation runs in the main paper use seed 42.
To assess reproducibility, we trained the best DARK configuration ($\beta_{\mathrm{start}}{=}2$, $\beta_{\mathrm{stop}}{=}{-}0.8$) with three different seeds.
\Cref{tab:seed_robustness} reports the results.

\begin{table}[tb]
  \caption{Seed robustness for DARK ($\beta_{\mathrm{start}}{=}2$, $\beta_{\mathrm{stop}}{=}{-}0.8$). All three seeds surpass the Static KD baseline (Composite Score = 0.826) and the FetalCLIP teacher (Composite Score =  0.853).}
  \label{tab:seed_robustness}
  \centering
  \begin{tabular}{@{}lcccc@{}}
    \toprule
    Seed & \fplane & \fbrain & \vrhc & Composite Score \\
    \midrule
    \textbf{42} & \textbf{0.946} & \textbf{0.784} & \textbf{0.886} & \textbf{0.886} \\
    123 & 0.945 & 0.656 & 0.882 & 0.860 \\
    7 & 0.933 & 0.764 & 0.845 & 0.857 \\
    \midrule
    Mean $\pm$ Std & $0.941 \pm 0.007$ & $0.735 \pm 0.069$ & $0.871 \pm 0.022$ & $0.867 \pm 0.016$ \\
    \bottomrule
  \end{tabular}
\end{table}

\noindent
Mean $\pm$ standard deviation across seeds: Composite Score $0.867 {\pm} 0.016$, \vrhc $0.871 {\pm} 2.2$\%.
\fbrain exhibits the highest variance ($0.735 {\pm} 0.069$), consistent with the greater difficulty of fine-grained brain sub-plane discrimination under a 26$\times$ teacher--student capacity gap. In contrast, \fplane ($0.941 {\pm} 0.007$) and \vrhc ($0.871 {\pm} 2.2$\%) are stable across all seeds. Critically, all three seeds surpass both the static KD baseline (Composite Score 0.826) and the FetalCLIP teacher (Composite Score 0.853), indicating that the overall gains from DARK are robust to random initialisation even when individual sub-metrics vary across seeds.

%% file: table_linear_probe.tex
\begin{table}[tb]
  \caption{%
    Linear probing evaluation.
    Frozen encoder features + single linear layer.
    95\% CIs from 5-fold $\times$ 5 seeds.
    \mfc retains 97--98\% of the teacher's performance at 26$\times$ fewer visual parameters.
  }
  \label{tab:linear_probe}
  \centering
  \scriptsize
  \setlength{\tabcolsep}{2.5pt}
  \begin{tabular}{@{}lccc@{}}
    \toprule
    Model & 6-View (F1) & Brain (F1) & CHD (AUROC) \\
    \midrule
    CLIP (ViT-L/14) & .867 {\tiny(.866--.869)} & .634 {\tiny(.632--.637)} & .679 {\tiny(.650--.708)} \\
    BiomedCLIP (ViT-B/16) & .856 {\tiny(.855--.858)} & .582 {\tiny(.577--.588)} & .643 {\tiny(.622--.665)} \\
    UniMed-CLIP (ViT-B/16) & .860 {\tiny(.858--.861)} & .607 {\tiny(.603--.610)} & .718 {\tiny(.702--.734)} \\
    FetalCLIP (ViT-L/14) & \textbf{.947} {\tiny(.947--.948)} & \textbf{.820} {\tiny(.818--.822)} & \textbf{.787} {\tiny(.770--.804)} \\
    \midrule
    \mfc (FastViT) & .930 {\tiny(.930--.931)} & .799 {\tiny(.797--.800)} & .769 {\tiny(.758--.779)} \\
    \quad \textit{Retention} & \textit{98.2\%} & \textit{97.4\%} & \textit{97.7\%} \\
    \bottomrule
  \end{tabular}
\end{table}

%% file: checklist.tex
\section*{NeurIPS Paper Checklist}
\label{checklist}

\begin{enumerate}

\item {\bf Claims}
    \item[] Question: Do the main claims made in the abstract and introduction accurately reflect the paper's contributions and scope?
    \item[] Answer: \answerYes{} 
    \item[] Justification: We clearly state our contributions and research scope in the claims presented in the abstract and introduction, all of which are supported by extensive experiments.

\item {\bf Limitations}
    \item[] Question: Does the paper discuss the limitations of the work performed by the authors?
    \item[] Answer: \answerYes{}
    \item[] Justification: We present the limitations of our paper in the section on Experiments and Results.

\item {\bf Theory assumptions and proofs}
    \item[] Question: For each theoretical result, does the paper provide the full set of assumptions and a complete (and correct) proof?
    \item[] Answer: \answerNA{} 
    \item[] Justification:  The paper does not include theoretical results. 

    \item {\bf Experimental result reproducibility}
    \item[] Question: Does the paper fully disclose all the information needed to reproduce the main experimental results of the paper to the extent that it affects the main claims and/or conclusions of the paper (regardless of whether the code and data are provided or not)?
    \item[] Answer: \answerYes{} 
    \item[] Justification:  Detailed information on experiments is provided in the Experimental section and the appendix.

\item {\bf Open access to data and code}
    \item[] Question: Does the paper provide open access to the data and code, with sufficient instructions to faithfully reproduce the main experimental results, as described in supplemental material?
    \item[] Answer: \answerNo{} 
    \item[] Justification: We submit our code in the supplementary code. Due to regulations and privacy constraints, the private training dataset cannot be released to the public.

\item {\bf Experimental setting/details}
    \item[] Question: Does the paper specify all the training and test details (e.g., data splits, hyperparameters, how they were chosen, type of optimizer) necessary to understand the results?
    \item[] Answer: \answerYes{}{} 
    \item[] Justification: We provide exhaustive training and test details in the paper.

\item {\bf Experiment statistical significance}
    \item[] Question: Does the paper report error bars suitably and correctly defined or other appropriate information about the statistical significance of the experiments?
    \item[] Answer: \answerYes{} 
    \item[] Justification: We report results for different seeds and also report mean and standard deviation.

\item {\bf Experiments compute resources}
    \item[] Question: For each experiment, does the paper provide sufficient information on the computer resources (type of compute workers, memory, time of execution) needed to reproduce the experiments?
    \item[] Answer: \answerYes{} 
    \item[] Justification: We report all details in the supplementary material.
    
\item {\bf Code of ethics}
    \item[] Question: Does the research conducted in the paper conform, in every respect, with the NeurIPS Code of Ethics \url{https://neurips.cc/public/EthicsGuidelines}?
    \item[] Answer: \answerYes{} 
    \item[] Justification: The research conform every respect of the NeurIPS Code of Ethics.

\item {\bf Broader impacts}
    \item[] Question: Does the paper discuss both potential positive societal impacts and negative societal impacts of the work performed?
    \item[] Answer: \answerYes{} 
    \item[] Justification: The paper discusses positive societal impact, such as enabling real-time fetal ultrasound AI on mobile or POCUS devices, while acknowledging the limitation that prospective validation is needed to ensure robustness before deployment.
    
\item {\bf Safeguards}
    \item[] Question: Does the paper describe safeguards that have been put in place for responsible release of data or models that have a high risk for misuse (e.g., pre-trained language models, image generators, or scraped datasets)?
    \item[] Answer: \answerNA{} 
    \item[] Justification: The paper poses no such risks.

\item {\bf Licenses for existing assets}
    \item[] Question: Are the creators or original owners of assets (e.g., code, data, models), used in the paper, properly credited and are the license and terms of use explicitly mentioned and properly respected?
    \item[] Answer: \answerYes{} 
    \item[] Justification: The license and terms of use are properly respected.

\item {\bf New assets}
    \item[] Question: Are new assets introduced in the paper well documented and is the documentation provided alongside the assets?
    \item[] Answer: \answerYes{}{} 
    \item[] Justification: We will release weights of distilled student model (MobileFetalCLIP).

\item {\bf Crowdsourcing and research with human subjects}
    \item[] Question: For crowdsourcing experiments and research with human subjects, does the paper include the full text of instructions given to participants and screenshots, if applicable, as well as details about compensation (if any)? 
    \item[] Answer: \answerNA{} 
    \item[] Justification: No crowdsourcing nor research with human subjects.

\item {\bf Institutional review board (IRB) approvals or equivalent for research with human subjects}
    \item[] Question: Does the paper describe potential risks incurred by study participants, whether such risks were disclosed to the subjects, and whether Institutional Review Board (IRB) approvals (or an equivalent approval/review based on the requirements of your country or institution) were obtained?
    \item[] Answer: \answerNA{} 
    \item[] Justification:  No crowdsourcing nor research with human subjects.

\item {\bf Declaration of LLM usage}
    \item[] Question: Does the paper describe the usage of LLMs if it is an important, original, or non-standard component of the core methods in this research? Note that if the LLM is used only for writing, editing, or formatting purposes and does \emph{not} impact the core methodology, scientific rigor, or originality of the research, declaration is not required.
    \item[] Answer: \answerNA{} 
    \item[] Justification: We do not utilize LLMs for our core method development.

\end{enumerate}